\documentclass{article}

\usepackage{microtype}
\usepackage{graphicx}
\usepackage{subcaption}
\usepackage{booktabs} 

\usepackage{hyperref}

\usepackage[preprint]{icml2026}

\usepackage{amsmath}
\usepackage{amssymb}
\usepackage{mathtools}
\usepackage{amsthm}

\usepackage{url}
\usepackage{amsfonts}       
\usepackage{nicefrac}       
\usepackage{xcolor}         
\usepackage[table]{xcolor}         
\usepackage{multirow}
\usepackage{multicol}
\usepackage{adjustbox}
\usepackage{subcaption}    
\usepackage{xspace}
\usepackage{dsfont}
\usepackage{wrapfig}

\definecolor{rank5}{HTML}{E0EDF9}
\definecolor{rank4}{HTML}{C8DFF5}
\definecolor{rank3}{HTML}{A2CBEE}
\definecolor{rank2}{HTML}{75AEE5}
\definecolor{rank1}{HTML}{5592DC}
\definecolor{bg}{HTML}{D9D9D9}
\definecolor{softgreen}{RGB}{34,139,34}

\newcommand{\eg}{\emph{e.g.,}\xspace}

\newcommand{\model}{CoME\xspace}
\newcommand{\hcr}{hybrid-capabilities reasoning\xspace}

\usepackage[capitalize,noabbrev]{cleveref}

\theoremstyle{plain}

\theoremstyle{definition}

\theoremstyle{remark}

\usepackage[textsize=tiny]{todonotes}

\icmltitlerunning{Submission and Formatting Instructions for ICML 2026}

\begin{document}

\twocolumn[
  \icmltitle{\model: Empowering Channel-of-Mobile-Experts with Informative Hybrid-Capabilities Reasoning}



  \icmlsetsymbol{intern}{$\dag$}
  \icmlsetsymbol{corresponding}{$\ddagger$}

  \begin{icmlauthorlist}
    \icmlauthor{Yuxuan Liu}{ruc,xiaomi,intern}
    \icmlauthor{Weikai Xu}{ntu,xiaomi,intern}
    \icmlauthor{Kun Huang}{xiaomi}
    \icmlauthor{Changyu Chen}{ruc,xiaomi,intern}
    \icmlauthor{Jiankun Zhao}{xiaomi}
    \icmlauthor{Pengzhi Gao}{xiaomi}
    \icmlauthor{Wei Liu}{xiaomi}
    \icmlauthor{Jian Luan}{xiaomi}
    \icmlauthor{Shang Shuo}{}
    \icmlauthor{Bo Du}{whu}
    \icmlauthor{Ji-Rong Wen}{ruc,corresponding}
    \icmlauthor{Rui Yan}{whu,corresponding}
  \end{icmlauthorlist}

  \icmlaffiliation{ruc}{Gaoling School of Artificial Intelligence, Renmin University of China}
  \icmlaffiliation{xiaomi}{MiLM Plus, Xiaomi Inc.}
  \icmlaffiliation{ntu}{Nanyang Technological University}
  \icmlaffiliation{whu}{Wuhan University}

  \icmlcorrespondingauthor{Ji-Rong Wen}{jrwen@ruc.edu.cn}
  \icmlcorrespondingauthor{Rui Yan}{rui.yan@whu.edu.cn}

  \icmlkeywords{Machine Learning, ICML}

  \vskip 0.3in
]



\printAffiliationsAndNotice{$^{\dag}$\ Work was done during the internship at XiaoMi.}  

\begin{abstract}
Mobile Agents can autonomously execute user instructions, which requires \hcr, including screen summary, subtask planning, action decision and action function. However, existing agents struggle to achieve both decoupled enhancement and balanced integration of these capabilities. 
To address these challenges, we propose \underline{\textbf{C}}hannel-\underline{\textbf{o}}f-\underline{\textbf{M}}obile-\underline{\textbf{E}}xperts (\textbf{\model}), a novel agent architecture consisting of four distinct experts, each aligned with a specific reasoning stage, \model activates the corresponding expert to generate output tokens in each reasoning stage via \textit{output-oriented activation}. 
To empower \model with \hcr, we introduce a progressive training strategy:
\textbf{Expert-FT} enables decoupling and enhancement of different experts' capability; \textbf{Router-FT} aligns expert activation with the different reasoning stage; \textbf{CoT-FT} facilitates seamless collaboration and balanced optimization across multiple capabilities.
To mitigate error propagation in \hcr, we propose InfoGain-Driven DPO (\textbf{Info-DPO}), which uses information gain to evaluate the contribution of each intermediate step, thereby guiding \model toward more informative reasoning.
Comprehensive experiments show that \model outperforms dense mobile agents and MoE methods on both AITZ and AMEX datasets.

\end{abstract}

\section{Introduction}

Mobile Agents can autonomously execute user instructions and have become a prominent research focus in both academia and industry. Their development is characterized by three major trends:
(1) from \textit{API invocation}~\cite{wen2023empowering, deng2024mobile} to \textit{action simulation}~\cite{xu2026mobilebenchv2realisticcomprehensivebenchmark, li2021learning}, enabling adaptation to more complex environment;
(2) from \textit{interactive exploration}~\cite{lee2023explore,wen2023droidbot} to \textit{supervised finetuning}~\cite{zhang2024you,cheng2024seeclick}, allowing to solve more generalized instructions; 
(3) from \textit{modular framework}~\cite{wang2024mobile,zhang2025appagent,li2024appagent} to \textit{holistic agent}~\cite{chai2024amex,lin2024showui,gou2024navigating}, simplifying system design and training pipelines.
Powered by advanced Multi-modal Large Language Models (MLLMs), mobile agents are undergoing a new paradigm shift from \textit{end-to-end prediction} to \textit{step-by-step reasoning}, which improves the accuracy and robustness of action decision~\cite{wu2024atlas,xu2024aguvis,qin2025ui}.

\begin{figure*}[]
\setlength{\belowcaptionskip}{-0.5cm}
\centering
\includegraphics[width=\linewidth]{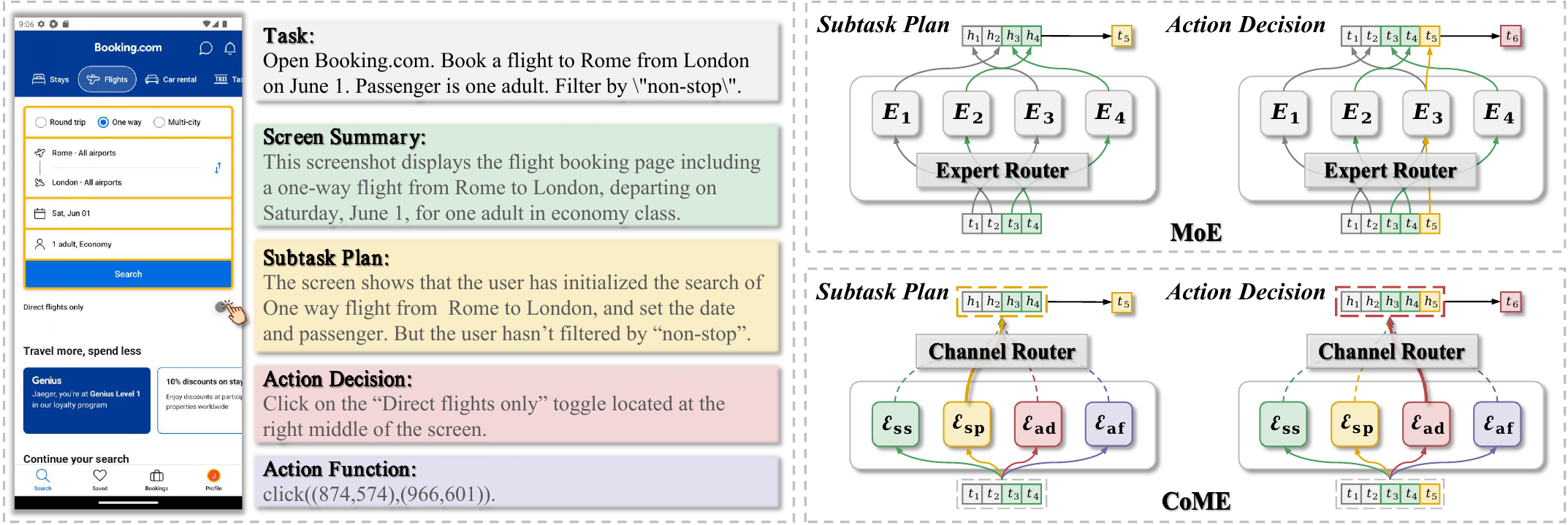}
\caption{Left is an example of \hcr. Right is the difference between input-oriented activation in MoE and output-oriented activation in \model.}
\label{fig:intro}
\end{figure*}

Achieving accurate action decisions remains challenging, even with Chain-of-Thought (CoT) prompting~\cite{wei2022chain}, as the reasoning process typically requires the agent to: perceive the current screen state, plan the next sub-task, generate high-level action decision and low-level action function~\cite{zhang2024android}. This process requires multi-dimensional capabilities, which is referred to as \textbf{\hcr} in Figure~\ref{fig:intro}. 
However existing mobile agents either enhance individual capabilities on task-specific datasets (\eg screen understanding~\cite{zhang2021screen,wang2021screen2words,li2021screen2vec} or action grounding~\cite{chai2024amex,gou2024navigating}), often lacking capabilities integration; or pre-train on large-scale dataset~\cite{cheng2024seeclick,wu2024atlas,qin2025ui}, which can lead to unbalanced performance of different capabilities. Effective methods for decoupled enhancement and balanced integration of multiple capabilities are still lacking.
Although Mixture-of-Experts (MoE) achieves partial capability decoupling by using input-oriented activation that routes different input tokens to different experts~\cite{zhou2022mixture,jiang2024mixtral,jamba,qwen_moe,dai2024deepseekmoe}, the ideal \hcr requires expert activation to align with the capabilities demanded to generate output tokens at each reasoning stage. Such output-oriented activation, however, is incompatible with the design of MoE.

To address these challenges, we propose a novel agent architecture: \underline{\textbf{C}}hannel-\underline{\textbf{o}}f-\underline{\textbf{M}}obile-\underline{\textbf{E}}xperts (\textbf{\model}), which incorporates four distinct experts to decouple \hcr, each specialized in one of the capabilities: screen summary, subtask plan, action decision, and action function. In contrast to the input-oriented activation in MoE, \model adopts output-oriented activation as shown in Figure~\ref{fig:intro}. 
Specifically, \model forwards the input tokens into each expert and selects the hidden states from the expert aligned with the current reasoning stage to generate each output token. 
To facilitate \model with \hcr, we design a progressive training strategy:
(1) \textbf{Expert-FT} trains FFN layers on capability-specific data to initialize each expert, achieving effective decoupling and enhancement of different capabilities;
(2) \textbf{Router-FT} trains channel router using expert label of each output token to align expert activation with reasoning stage;
(3) \textbf{CoT-FT} trains \model with the \hcr data to enable seamless collaboration and balanced optimization among different experts.
Through this progressive curriculum, \model achieves \hcr by activating specific expert aligned with reasoning stages.

As \hcr spans multiple stages, even minor errors in intermediate steps can propagate and compromise the final outcome. 
To address this, we propose \textbf{InfoGain-Driven DPO (Info-DPO)}, which leverages information gain to quantify the contribution of each intermediate step to the final action prediction.
Specifically, we use reward model to estimate the information entropy to the ground truth action before and after each reasoning stage, and use the reduction in entropy as InfoGain reward.
Combined with action accuracy reward, we distinguish effective reasoning trajectories to construct high-quality DPO data.
Through this mechanism, Info-DPO encourages the model to reinforce informative and reliable intermediate steps, while suppressing those contain distraction reasoning steps. Consequently, the model improves the reasoning accuracy at each stage and mitigates the error propagation throughout the reasoning trajectory.
Experiments on the AITZ and AMEX datasets demonstrate that \model outperforms dense mobile agents (+1.73\%) and sparse MoE models (+5.72\%) with equivalent activated parameters. Further analysis verifies that \model achieves accurate expert activation aligned with reasoning stage, while Info-DPO improves the effectiveness of intermediate reasoning steps.
Overall, our main contributions can be summarized as follows:

$\bullet$ We propose Channel-of-Mobile-Experts (\textbf{\model}), a novel architecture incorporates experts specialized in: screen summary, subtask plan, action decision and action function. \model employs output-oriented activation to activate appropriate expert at each stage of \hcr.

$\bullet$ We develop a progressive training strategy to empower \model for \hcr: \textbf{Expert-FT} enables decoupling and enhancement of different capabilities to profile specific experts; \textbf{Router-FT} aligns expert activation with the reasoning stage; \textbf{CoT-FT} facilitates seamless collaboration and balanced optimization among experts.

$\bullet$ We introduce InfoGain-Driven DPO (\textbf{Info-DPO}), which measures the contribution of intermediate reasoning steps via information gain to suppress invalid reasoning and mitigate error propagation in DPO training. Comprehensive experiments demonstrate the effectiveness \model.

\section{Related works}

\subsection{Autonomous mobile agents}

Mobile Agents can autonomously execute user instruction through API invocation or action simulation~\cite{wen2023empowering,deng2024mobile,li2020mapping,li2021learning}, which have become a pivotal research spotlight. 
To equip the agents with mobile knowledge, previous approaches introduce a prior exploration stage~\cite{lee2023explore, wen2023droidbot, li2024appagent}, or design multiple specific tasks, such as screen understand~\cite{zhang2021screen,wang2021screen2words,li2021screen2vec}, widgets recognition~\cite{chen2022towards,li2020widget,zheng2024agentstudio}, GUI transition~\cite{gou2024navigating,wu2024mobilevlm} and element/action grounding~\cite{cheng2024seeclick,chai2024amex,bai2021uibert,baechler2024screenai,qian2024visual}.
Empowered by the advanced MLLMs, some research design some framework consists of multiple stages or agents~\cite{wang2024mobile,li2024appagent,wang2025mobile,liu2025mobilesteward}.
Recently, more works pre-train or finetune on large scale of mixed data to build general mobile agents~\cite{zhang2024you,huang2025mobileiplenhancingmobileagents,xu2024aguvis,qin2025ui,chen2025stepsuccessrateawaretrajectoryefficientpolicy}.
While existing mobile agents still struggle in capability disentanglement and balanced integration. Therefore, we propose Channel-of-Mobile-Experts (\model) to facilitate \hcr.

\subsection{Mixture-of-Experts}

Mixture‐of‐Experts (MoE)~\cite{jacobs1991adaptive} integrates multiple experts and dynamically routes each input to the most relevant expert, to address diverse tasks~\cite{shazeer2017outrageously,lepikhingshard,fedus2022switch}. 
Recent works have integrated MoE into LLMs~\cite{jiang2024mixtral,jamba,qwen_moe,dai2024deepseekmoe} and MLLMs~\cite{li2025uni,deitke2024molmo,wu2024deepseek} to boost capacity and efficiency by extending FFN layers into multiple experts and activating only top-K experts for each input token. 
AriaUI~\cite{yang2024aria}, the first MoE GUI agent, demonstrates the potential of MoE for mobile automation. 
However, MoE relies on input-oriented activation, which is not optimal for \hcr. In contrast, \model employs output-oriented activation to align expert activation with the reasoning stage.

\section{Methodology}

\subsection{Task formulation}

Given a user instruction $I$, at each step, the \textbf{Mobile Agent} $\mathcal{M}$ performs \textbf{\hcr} on the current screen state $S$ and action history $H$ to generate the next action $a$. The reasoning trajectory $T$ comprises four distinct stages: screen summary ($T_{\text{ss}}$), subtask ($T_{\text{sp}}$), action decision ($T_{\text{ad}}$), and action function invocation ($T_{\text{af}}$). The action $a$ is extracted from $T_{\text{af}}$.

\vspace{-5pt}
\begin{equation}
a,\ T = \mathcal{M}(I, S, H), \ \text{where}\ T = \left[ T_{\text{ss}},T_{\text{sp}},T_{\text{ad}},T_{\text{af}} \right].
\end{equation}

\subsection{Channel-of-Mobile-Experts (\model)}

\begin{figure}[]
  \setlength{\abovecaptionskip}{0.2cm}
  \setlength{\belowcaptionskip}{-0.5cm}
  \centering
  \includegraphics[width=0.9\linewidth]{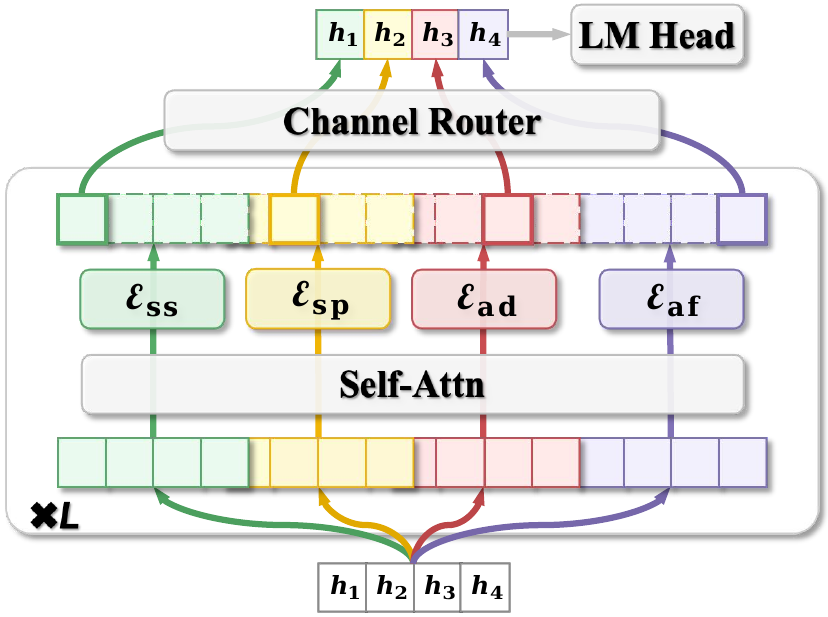}
  \caption{\model architecture.}
  \label{fig:expert_performance}
\end{figure}
Mobile agents can automate user instructions through careful \hcr, but still face two main challenges: 
dense models struggle to achieve decoupled enhancement and balanced integration of different capabilities, while MoE models fail to activate the expert aligned with the reasoning stage.

To address these challenges, we propose Channel‑of‑Mobile-Experts (\textbf{\model}), a novel agent architecture with \textit{output‑oriented activation}. 
\model extends FFN module in each layer with four different experts  $\bigl[\mathcal{E}_{\text{ss}},\mathcal{E}_{\text{sp}},\mathcal{E}_{\text{ad}},\mathcal{E}_{\text{af}}\bigr]$, while shares the same Self-Attn module among experts. 
In contrast to the input-oriented activation in MoE\footnote{Limitation of MoE is analysed in Appendix~\ref{appendix:moe}}, \model adopts output-oriented activation to activates expert with the capability required by the current reasoning stage.
Specifically, given the hidden states $\mathbb{H}\in\mathbb{R}^{B\times N \times D}$ after the embedding layer, where $B$, $N$, $D$ stands for the batch size, sequence length and hidden size. \model first repeats the hidden states for $E$ times $\mathbb{H} \rightarrow \mathbb{\widetilde{H}}\in\mathbb{R}^{B\times N \times E \times  D}$, where $E$ represents the number of experts and $\mathbb{\widetilde{H}}$ represents the channel hidden states.
At the $l$-th layer in \model:

\vspace{-10pt}
\begin{equation}
\begin{aligned}
     \mathbb{\widetilde{H}}^{(l)} &= \operatorname{Self-Attn}\,\bigl(\mathbb{\widetilde{H}}^{(l-1)}\bigr) \\ 
     \mathbb{\widetilde{H}}^{(l)}_{[e]} &= \operatorname{FFN}_e\,\bigl(\mathbb{\widetilde{H}}^{(l)}_{[e]}\bigr), \quad e \in[1,\cdots,E]
\end{aligned}
\end{equation}
\vspace{-5pt}

After obtaining the channel hidden state $\mathbb{\widetilde{H}}^{(L)}$ from the last layer, the most important step in \model to achieve output-oriented activation is to select the hidden state from the corresponding expert channel according to the current reasoning stage. The channel router $\mathbf{W}_c \in \mathbb{R}^{ED \times E}$ will project the flattened hidden states into $E$-dimension channel logits $\mathbb{C} \in \mathbb{R}^{B\times N \times E}$, which will be used to fuse the hidden states from different expert channels to generate the final hidden states $\mathbb{\widehat{H}} \in \mathbb{R}^{B\times N \times D}$, achieving output-oriented activation. Then $\mathbb{\widehat{H}}$ will be forwarded into $\text{LM-Head}$.

\vspace{-10pt}
\begin{equation}
\begin{aligned}
\mathbb{C} &= \operatorname{flatten}\bigl(\mathbb{\widetilde{H}}^{(L)}\bigr)\times\mathbf{W}_c,\\
\widehat{\mathbb{{H}}}_{[b,n]} &= 
    \sum_{e=1}^{E}
    \operatorname{softmax}\bigl(\mathbb{C}\bigr)_{[e]} \cdot \mathbb{\widetilde{H}}^{(L)}_{[e]}, 
\end{aligned}
\end{equation}
\vspace{-15pt}

\subsection{Progressive training strategy}

In order to empower \model with \hcr, we introduce a progressive training strategy: 
(1) \textbf{Expert Finetuning (Expert-FT)}, which explicitly decouples and enhances different capabilities; 
(2) \textbf{Router Finetuning (Router-FT)}, which allows expert activation aligned with the current reasoning stage; 
(3) \textbf{Chain-of-Thought Finetuning (CoT-FT)}, which facilitates seamless collaboration and balanced optimization among experts. 
Details in the following sections.

\subsubsection{Stage-1: expert finetuning (Expert-FT)}

The \hcr can be divided into four stages: screen summary, subtask plan, action decision and action function~\cite{zhang2024android}, which is explicitly decoupled with different experts in \model. To initialize and enhance these specialized experts, we train the FFN modules in Qwen2VL~\cite{wang2024qwen2vl} dense model on the dataset $\mathcal{D}_e$ of a specific ability respectively to initialize the different expert $e \in \bigl[\mathcal{E}_{\text{ss}},\mathcal{E}_{\text{sp}},\mathcal{E}_{\text{ad}},\mathcal{E}_{\text{af}}\bigr]$

\vspace{-6pt}
\begin{equation}
    \mathcal{L}_{\text{Expert-FT}} = -\mathbb{E}_{\{x,y\} \sim\mathcal{D}_e}\bigl[\log \pi_{\mathcal{M}_e}\bigl(y \,|\,x\bigr)\bigr]
\end{equation}

We extract the FFN layers in the specialized experts $\bigl[\mathcal{E}_{\text{ss}},\mathcal{E}_{\text{sp}},\mathcal{E}_{\text{ad}},\mathcal{E}_{\text{af}}\bigr]$ to assemble \model. 

\vspace{-6pt}
\begin{equation}
    \text{FFN}_{\text{\model}}^{(l)} = \Bigl[\text{FFN}_{\text{ss}}^{(l)}\,,\,\text{FFN}_{\text{sp}}^{(l)}\,,\,\text{FFN}_{\text{ad}}^{(l)}\,,\,\text{FFN}_{\text{af}}^{(l)} \Bigr] \notag
\end{equation}
\vspace{-15pt}

\subsubsection{Stage-2: router finetuning (Router-FT)}

After the Expert-FT, \model has potentially mastered the capabilities required by \hcr, thus it is necessary to enable output-oriented activation in \model to align expert activation with the reasoning stage. We augment the \hcr data $\mathcal{D}$ with the activated expert label of each output token according to the reasoning stage it is in.
During training, we optimize the channel router using the Cross-Entropy loss $\mathcal{L}_{\text{R-CE}}$ between the predicted channel logits $\mathbb{C} \in \mathbb{R}^{B\times N\times E}$ and the expert labels $\mathcal{C} \in \mathbb{R}^{B\times N}$. To further supervise expert activation, we apply a Router Norm loss $\mathcal{L}_{\text{R-Norm}}$ as a regularization term to suppress irrelevant experts.

\vspace{-6pt}
{\fontsize{9.5pt}{11pt}\selectfont
\begin{equation}
\begin{gathered}
\mathcal{L}_{\text{R-CE}} = -\mathbb{E}_{\{\mathcal{C},\mathbb{C}\} \sim\mathcal{D}}
\bigl[
\mathcal{C}\cdot 
\log\mathop{\mathrm{softmax}}\bigl(\mathbb{C}\bigr) 
\bigr] \\
\mathcal{L}_{\text{R-Norm}} = -\mathbb{E}_{\{\mathcal{C},\mathbb{C}\} \sim\mathcal{D}}
\bigl[
\lVert \operatorname{softmax}\,\bigl(\mathbb{C}\bigr),
\operatorname{onehot}\,\bigl(\mathcal{C}\bigr) \Vert_2^2
\bigr] \\
\mathcal{L}_{\text{Router-FT}} = \mathcal{L}_{\text{R-CE}} + \mathcal{L}_{\text{R-Norm}}
\end{gathered} 
\end{equation}
}

\subsubsection{Stage-3: Chain-of-Thought finetuning (CoT-FT)}

\model achieves the decoupling and enhancement of multi-dimensional capabilities through Expert-FT, and aligns the expert activation with the reasoning stage through Router-FT. 
To further improve \hcr, we design CoT Finetuning, which aims to simultaneously facilitates seamless collaboration and balanced optimization among different experts. During training, we use SFT loss $\mathcal{L}_{\text{SFT}}$ to optimize the hybrid-capabilities reasoning, and use a regularization term $\mathcal{L}_{\text{R-Norm}}$ to constrain expert activation during reasoning. We use $X$ to denote $\{I,S,H\}$.

\vspace{-6pt}
\begin{equation}
\begin{gathered}
    \mathcal{L}_{\text{SFT}} = -\mathbb{E}_{\{X,T\} \sim\mathcal{D}}\bigl[\log \pi_{\mathcal{M}}\bigl(T\ |\ X\bigr)\bigr] \\
    \mathcal{L}_{\text{CoT-FT}} = \mathcal{L}_{\text{SFT}} + \gamma \cdot \mathcal{L}_{\text{R-Norm}}
\end{gathered}
\label{eq:sft_loss}
\end{equation}

\subsection{InfoGain-driven DPO}

\begin{figure}[]
  \centering
  \includegraphics[width=0.45\textwidth]{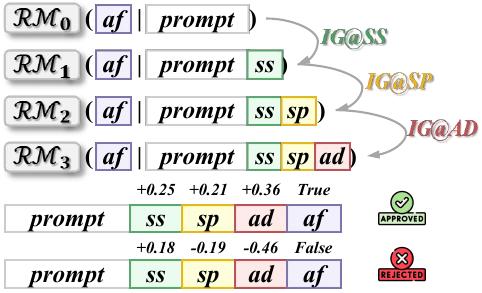}
  \caption{An example of InfoGain reward.}
  \label{fig:reward}
  \vspace{-10pt}
\end{figure}
Chain-of-Thought stimulates the model's reasoning ability through step-by-step thinking to get more accurate results~\cite{wei2022chain}. However, the intermediate reasoning steps may contain errors, which will impact the final accuracy because of the error propagation as the CoT length increases. 
In addition, reasoning trajectories that reach the correct answer via flawed steps should be suppressed, while those with logical steps but incorrect outcomes may still be beneficial. Therefore the key to improve the accuracy of reasoning is to ensure that each intermediate step has a positive contribution to the final answer. Since \hcr is naturally a multi-stage reasoning structure, we can use the information gain of each intermediate stage to measure its contribution~\cite{ton2024understanding}.

\vspace{-10pt}
\begin{equation}
\small
    IG(T_{e}) = \log \frac{p\bigl(\,T_{\text{af}}\:|\:X,T_{[0:e]}\bigr)}{p\bigl(\,T_{\text{af}}\:|\:X,T_{[0:e-1]}\bigr)}
\label{eq:infogain-1}
\end{equation}

In order to estimate the information gain of the $e$-th reasoning stage, we train the reward model $\mathcal{RM}_e$, using the reasoning stages $T_{[0:e]}$ as additional inputs to directly predict the final action function $T_{\text{af}}$.

\vspace{-10pt}
{\fontsize{8.5pt}{11pt}\selectfont
\begin{equation}
    \mathcal{L}_{\mathcal{RM}_e} = -\mathbb{E}_{\{X,T\} \sim\mathcal{D}}\bigl[\log \pi_{\mathcal{RM}_e}\bigl(T_\text{af}\ |\ X,T_{[0:e]}\bigr)\bigr]
\end{equation}
}

\begin{table*}[]
\renewcommand\arraystretch{1.1}
\centering
\caption{\textbf{Main results on AITZ dataset of different action.} Results with $^*$ are reported from ~\cite{zhang2024android}. Methods with $^\dagger$ are finetuned with \hcr. The best overall result is marked \textbf{bold} and the second-best one is marked \underline{underline}.}
\resizebox{\textwidth}{!}{
\begin{tabular}{lcccccccccc}
\toprule
\midrule
\multirow{2}{*}{\textbf{Method}} & \multirow{2}{*}{\textbf{\#Params}} & \multirow{2}{*}{\textbf{SCROLL}} & \multicolumn{2}{c}{\textbf{CLICK}} & \multicolumn{2}{c}{\textbf{TYPE}} & \multirow{2}{*}{\textbf{PRESS}} & \multirow{2}{*}{\textbf{STOP}} & \multicolumn{2}{c}{\textbf{Overall}} \\ \cmidrule(lr){4-5} \cmidrule(lr){6-7} \cmidrule(lr){10-11}
    &  &   & \textbf{type}   & \textbf{match}   & \textbf{type}   & \textbf{match}  &  & & \textbf{type}   & \textbf{match}  \\ \midrule
Auto-GUI$^*$~\cite{zhang2024you}      & 700M & 61.40 & 74.56 & 32.20 & 87.80 & 81.40 & 57.70 & 74.40 & \textbf{82.98} & 47.69 \\
Qwen2VL$^\dagger$~\cite{wang2024qwen2vl}     & 2B   & 23.05 & 78.38 & 43.74 & 60.40 & 59.40 & 52.22 & 47.02 & 62.25 & 43.80 \\
ShowUI~\cite{lin2024showui}      & 2B   & 23.79 & 53.29 & 40.34 & 89.60 & 85.20 & 59.00 & 87.55 & 64.22 & 49.55 \\
UITars~\cite{qin2025ui}      & 2B   & 30.17 & 83.81 & 55.25 & 83.97 & 84.77 & 53.78 & 61.39 & 74.23 & 55.63 \\ 
GUI-R1~\cite{luo2025gui} & 3B & 25.53 & 79.00 & 50.47 & 21.80 & 20.80 & 32.89 & 28.77 & 57.32 & 40.41 \\
\midrule
SeeClick~\cite{cheng2024seeclick}    & 7B   & 11.14 & 69.92 & 52.96 & 53.80 & 53.00 & 67.88 & 55.36 & 62.93 & 49.11 \\
Qwen2VL$^\dagger$~\cite{wang2024qwen2vl}     & 7B   & 43.28 & 83.97 & 55.30 & 51.80 & 51.40 & 56.92 & 64.87 & 71.59 & 54.46 \\
UGround~\cite{gou2024navigating}    & 7B   & 58.22 & 80.94 & 58.48 & 82.56 & 73.85 & 58.22 & 68.78 & 74.54 & 60.19 \\
OS-Atlas~\cite{wu2024atlas}    & 7B   & 76.12 & 75.82 & 54.83 & 89.80 & 88.60 & 68.67 & 81.75 & 77.83 & 65.11 \\
UITars~\cite{qin2025ui}      & 7B   & 56.50 & 84.87 & 63.87 & 85.97 & 85.77 & 58.22 & 71.29 & 78.07 & \underline{65.41} \\ 
GUI-R1~\cite{luo2025gui} & 7B & 29.02 & 82.22 & 53.88 & 29.60 & 27.00 & 45.17 & 42.26 & 62.87 & 45.91 \\
UI-S1~\cite{lu2025ui} & 7B & 45.60 & 78.75 & 56.73 & 91.20 & 87.20 & 51.17 & 39.28 & 69.85 & 56.22 \\
\midrule
MolmoE$^\dagger$~\cite{deitke2024molmo}      & 1B   & 28.19 & 79.84 & 33.17 & 65.20 & 62.00 & 37.07 & 54.96 & 65.96 & 38.23 \\
Qwen2VL-MoE$^\dagger$ & 3B   & 48.75 & 80.03 & 59.43 & 77.80 & 74.00 & 57.70 & 70.83 & 72.94 & 60.69 \\
AriaUI~\cite{yang2024aria}      & 3.9B & 53.73 & 85.51 & 60.20 & 84.20 & 80.80 & 63.70 & 76.38 & 78.53 & 63.56 \\
DeepSeekVL2$^\dagger$~\cite{wu2024deepseek} & 4.5B & 17.94 & 75.35 & 19.98 & 50.90 & 46.69 & 14.36 & 24.25 & 55.36 & 22.55 \\ \midrule
\textbf{\model} & 5B   & 52.07 & 83.83 & 65.22 & 88.20 & 83.80 & 59.53 & 83.33 & \underline{78.60} & \textbf{66.98} \\
\midrule
\bottomrule
\end{tabular}
}
\label{tab:aitz}
\vspace{-12pt}
\end{table*}

Therefore, the information gain of the $e$-th reasoning stage in Eq~\ref{eq:infogain-1} can be approximated by Eq~\ref{eq:infogain}. Specifically, we denote the information gain in screen summary, subtask plan and action decision stages as \textit{IG@SS}, \textit{IG@SP} and \textit{IG@AD} respectively as shown in Figure~\ref{fig:reward}. Then we can calculate the reasoning-level InfoGain reward $\mathcal{R}_\text{IG} = \textit{IG@SS}+\textit{IG@AP}+\textit{IG@AD}$ and the InfoPos reward $\mathcal{R}_{\text{IG}}^+ = \mathds{1}_{\textit{IG@SS}>0} \cdot \mathds{1}_{\textit{IG@AP}>0} \cdot \mathds{1}_{\textit{IG@AD}>0}$.

\vspace{-5pt}
\begin{equation}
    IG(T_{e}) \approx
 \log \frac{\pi_{\mathcal{RM}_e}\bigl(\,T_{\text{af}}\:|\:X,T_{[0:e]}\bigr)}{\pi_{\mathcal{RM}_{e-1}}\bigl(\,T_{\text{af}}\:|\:X,T_{[0:e-1]}\bigr)}
\label{eq:infogain}
\end{equation}

We use the action‐level accuracy reward $\mathcal{R}_\text{ACC}$ 
in Eq~\ref{eq:action_acc}. For click actions, $\mathcal{R}_{\mathrm{ACC}}$ is computed based on the distance between the predicted and annotated coordinates within the threshold $\delta_d$; for input actions, it is the F1 score between the predicted and reference text within the threshold $\delta_f$; and for other actions, it is a binary exact‐match indicator.

\vspace{-10pt}
{\fontsize{9.5pt}{11pt}\selectfont
\begin{equation}
\begin{aligned}
    \mathcal{R}_\text{ACC} = 
    \begin{cases}
        1-\max(\operatorname{dis}(a,\hat{a})\,/\,\delta_d, 1), & \text{click action} \\
        \max(\operatorname{f1}(a,\hat{a}), 0) \cdot \mathds{1}_{ \operatorname{f1} > \delta_f }, & \text{input action} \\
        \mathds{1}_{ a = \hat{a} }, & \text{other actions}
    \end{cases}
\end{aligned}
\label{eq:action_acc}
\end{equation}
}

When generating DPO data, we sample $K$ reasoning trajectories for each action, and calculate the CoT-reward $\mathcal{R}_\text{CoT} = \mathcal{R}_\text{IG} \cdot \mathcal{R}_\text{ACC}$. We select the trajectory with the highest $\mathcal{R}_\text{CoT}$ as well as $\mathcal{R}_{\text{IG}}^+ =1$ from the trajectories that hit the labeled action as the \textit{chosen output}, which means that all of the reasoning stages are on the correct direction and lead to the most approximate answer to the labeled action. The \textit{rejected output} is selected from the trajectories that don't hit the ground truth action and has the lowest $\mathcal{R}_\text{IG}$. More details about the DPO data selection can be found in Appendix~\ref{appendix:dpo_pair}. After obtaining the DPO dataset $\mathcal{D}^*$, we can train \model using the following DPO loss together with SFT loss $\mathcal{L}_\text{SFT}$ and Router Norm loss $\mathcal{L}_\text{R-Norm}$:

\vspace{-15pt}
\begin{align}
\mathcal L_{\text{IG-DPO}}
&= - \mathbb E_{\{X,T^+,T^-\}\sim \mathcal{D}^*} 
\bigg[\log\sigma\!\big(\beta(\Delta_{\mathcal{M}}-\Delta_{\mathcal{M}_{\text{ref}}})\big)\bigg] \notag\\
&\quad + \alpha \cdot \mathcal{L}_\text{SFT} + \gamma \cdot \mathcal{L}_\text{R-Norm}
\label{eq:dpo_loss}\\
&\text{where}\quad
\Delta_{\theta} = \log\frac{\pi_{\theta}(T^+|X)}{\pi_{\theta}(T^-|X)}
\notag
\end{align}
\vspace{-15pt}

\section{Experiment}

\begin{table*}[t]
\renewcommand\arraystretch{1.1}
\centering
\caption{\textbf{Main results on AMEX dataset of different apps}. Results with $^*$ are reported from ~\cite{chai2024amex}. Methods with $^\dagger$ are finetuned with \hcr. The best overall result is marked \textbf{bold} and the second-best one is marked \underline{underline}.}
\resizebox{\textwidth}{!}{%
\begin{tabular}{lccccccccccc}
\toprule
\midrule
\textbf{Method} & \textbf{\#Params} & \textbf{Gmail} & \textbf{Booking} & \textbf{Music} & \textbf{SHEIN} & \textbf{News} & \textbf{CM}	& \textbf{ToDo}	&  \textbf{Signal}	&  \textbf{Yelp} & \textbf{Overall}\\
\midrule 
Qwen2VL$^\dagger$~\cite{wang2024qwen2vl}     & 2B   & 37.4 & 35.2 & 40.1 & 33.7 & 50.0 & 47.5 & 41.8 & 56.7 & 40.5 & 38.53 \\
ShowUI~\cite{lin2024showui}      & 2B   & 52.2 & 33.6 & 68.8 & 55.3 & 51.8 & 57.4 & 51.9 & 69.7 & 50.2 & 47.74 \\
UITars~\cite{qin2025ui}      & 2B   & 59.4 & 47.9 & 55.4 & 53.0 & 65.1 & 62.3 & 64.7 & 61.7 & 58.1 & 54.43 \\ 
GUI-R1~\cite{luo2025gui} & 3B & 37.1 & 50.6 & 58.2 & 47.1 & 21.6 & 54.7 & 54.6 &  56.1 & 54.4 & 48.53 \\
\midrule
SeeClick$^*$~\cite{cheng2024seeclick}    & 7B   & 28.2 & 29.4 & 18.1 & 20.0 & 30.0 & 53.1 & 30.7 & 37.1 & 27.4 & 30.44 \\
Qwen2VL$^\dagger$~\cite{wang2024qwen2vl}     & 7B   & 57.6 & 58.4 & 56.5 & 47.3 & 64.2 & 66.3 & 60.9 & 72.8 & 54.8 & 57.99 \\
UGround~\cite{gou2024navigating}    & 7B   & 70.9 & 68.8 & 72.7 & 63.7 & 77.7 & 67.7 & 73.7 & 80.1 & 67.6 & 69.12 \\
OS-Atlas~\cite{wu2024atlas}    & 7B   & 74.4 & 69.7 & 74.6 & 64.0 & 80.7 & 63.6 & 65.3 & 83.3 & 62.3 & 69.99 \\
SphAgent$^*$~\cite{chai2024amex}    & 7B   & 61.7 & 68.2 & 77.7 & 72.0 & 71.9 & 64.6 & 79.6 & 71.3 & 69.6 & \underline{70.71} \\
UITars~\cite{qin2025ui}      & 7B   & 67.7 & 70.0 & 71.8 & 63.8 & 71.5 & 67.7 & 77.0 & 86.4 & 72.8 & 70.33 \\ 
GUI-R1~\cite{luo2025gui} & 3B & 36.8 & 53.0 & 64.6 & 49.4 & 20.6 & 68.6 & 64.5 & 68.5 & 62.3 & 52.47 \\
UI-S1~\cite{lu2025ui} & 7B & 47.3 & 62.2 & 66.0 & 54.1 & 38.1 & 75.3 & 72.7 & 80.2 & 63.6 & 60.57 \\
\midrule
MolmoE$^\dagger$~\cite{deitke2024molmo}      & 1B   & 38.7 & 28.6 & 28.1 & 27.2 & 45.3 & 22.8 & 37.6 & 38.2 & 34.0 & 31.56 \\
Qwen2VL-MoE$^\dagger$ & 3B   & 65.1 & 63.7 & 57.8 & 63.0 & 78.0 & 60.3 & 70.2 & 76.5 & 61.1 & 64.56 \\
AriaUI~\cite{yang2024aria}      & 3.9B & 63.1 & 62.3 & 68.5 & 58.9 & 83.0 & 54.7 & 62.5 & 83.3 & 66.9 & 64.10 \\
DeepSeekVL2$^\dagger$~\cite{wu2024deepseek} & 4.5B & 43.1 & 36.9 & 52.5 & 42.2 & 42.7 & 51.1 & 47.0 & 61.7 & 38.9 & 42.22 \\ \midrule
\textbf{\model} & 5B   & 76.2 & 72.6 & 81.0 & 64.3 & 81.2 & 63.2 & 72.6 & 78.4 & 66.9 & \textbf{72.61} \\
\midrule
\bottomrule

\end{tabular}
}
\label{tab:amex}
\vspace{-12pt}
\end{table*}

\subsection{Experiment setup}

We train and evaluate \model on two datasets, \textbf{AITZ}~\cite{zhang2024android} and \textbf{AMEX}~\cite{chai2024amex}. We compare against 13 baselines, covering both dense mobile agents and sparse MoE models, and report the accuracy of action type and match. More details on datasets, baselines, metrics, and implementation are provided in the Appendix~\ref{appendix:setup}.

\subsection{Main results}

\noindent \textbf{AITZ.} As shown in Table~\ref{tab:aitz}, \model achieves the highest overall action match accuracy. Compared to dense mobile agents,\model yields an improvement of 11.35\% over 2B-series model and 1.57\% over 7B-series models, with only 5B activated parameters. Moreover, \model surpasses MoE-based models by 3.42\%.
Baselines generally underperform on the CLICK action, while \model achieves the highest accuracy of 65.22\% (+1.45\%), because CLICK is a representative hybrid-capabilities task that is much more challenging.
Baselines also show imbalanced performance across action types (\eg, ShowUI: 87.55\% STOP vs. 23.79\% SCROLL). In contrast, \model achieves a higher relative improvement (+11.56\%) and lower bias (4.41) in Table~\ref{tab:aitz-improve}, indicating more balanced action performance.

\begin{table}[]
\renewcommand\arraystretch{0.9}
\centering
\setlength{\abovecaptionskip}{0.3cm}
\caption{Ablation analysis on AITZ and AMEX.}
\resizebox{0.9\linewidth}{!}{%
\begin{tabular}{lccc}
\toprule
\midrule
\textbf{Method}& \textbf{AITZ}& \textbf{AMEX}& \textbf{AVG}\\ \midrule
\model & 66.98       & 72.61       & 69.78      \\ \midrule
- w/o Info-DPO    & 62.93       & 67.28       & 65.10      \\
- w/o Router-FT    & 60.05       & 62.00       & 61.02      \\
- w/o Expert-FT   & 57.07       & 64.47       & 60.74      \\ \midrule
- Info-DPO w/o $\mathcal{L}_\text{R-Norm}$   & 65.96       & 70.90       & 68.42      \\
- CoT-FT w/o $\mathcal{L}_\text{R-Norm}$   & 62.57       & 64.24       & 63.40      \\
\midrule
\bottomrule
\end{tabular}
}
\label{tab:ablation}
\vspace{-18pt}
\end{table}

\noindent \textbf{AMEX.} As shown in Table~\ref{tab:amex}, \model achieves the best overall performance across nine apps, surpassing the dense model (+1.90\%) and the sparse MoE (+8.05\%). 
Compared with OS-Atlas, SphAgent and UITars pre-trained with large scale mobile data then finetuned on AMEX, \model could surpasses them by 2.26\% on average, only using AMEX data through \hcr, which proves that \model can better activate multi-capabilities to achieve more effective CoT reasoning.

\subsection{Ablation analysis}

We designed comprehensive ablation experiments to analyze the effects of different training stages and strategies. 
As shown in Table~\ref{tab:ablation}, Info-DPO contributes most to the action prediction (+4.68\%), proving that using information gain to distinguish reasoning trajectories can mitigate error propagation and improve reasoning accuracy. 
Further, removing Router-FT leads to an accuracy decline (-4.08\%), indicating that Router-FT enables expert activation aligned with the reasoning stage. 
Moreover, no prior Expert-FT can not fully release the expert's specialized capability (-4.36\%).
Including router norm $\mathcal{L}_\text{R-Norm}$ on expert activation is necessary in CoT-FT (+1.70\%) and Info-DPO (+1.36\%) as well.
\begin{figure}[]
  \centering
  \includegraphics[width=\linewidth]{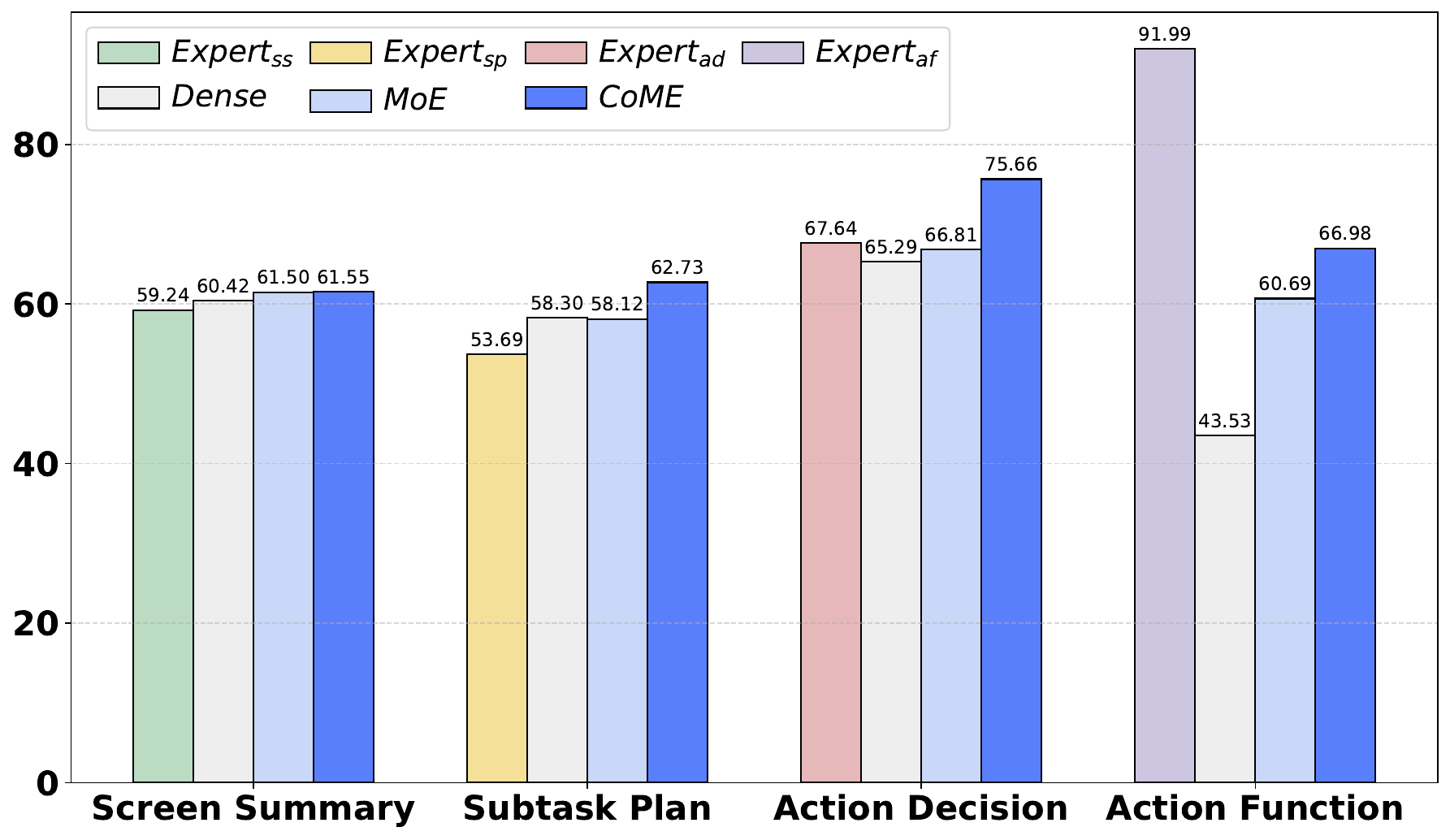}
  \caption{Performance on different reasoning stages.}
  \label{fig:expert_performance}
  \vspace{-15pt}
\end{figure}

\vspace{-5pt}
\subsection{Comprehensive analysis of \model design}

\begin{figure*}[]
\setlength{\abovecaptionskip}{5pt}
  \centering
  \begin{subfigure}[b]{0.49\textwidth}
    \centering
    \includegraphics[width=\linewidth]{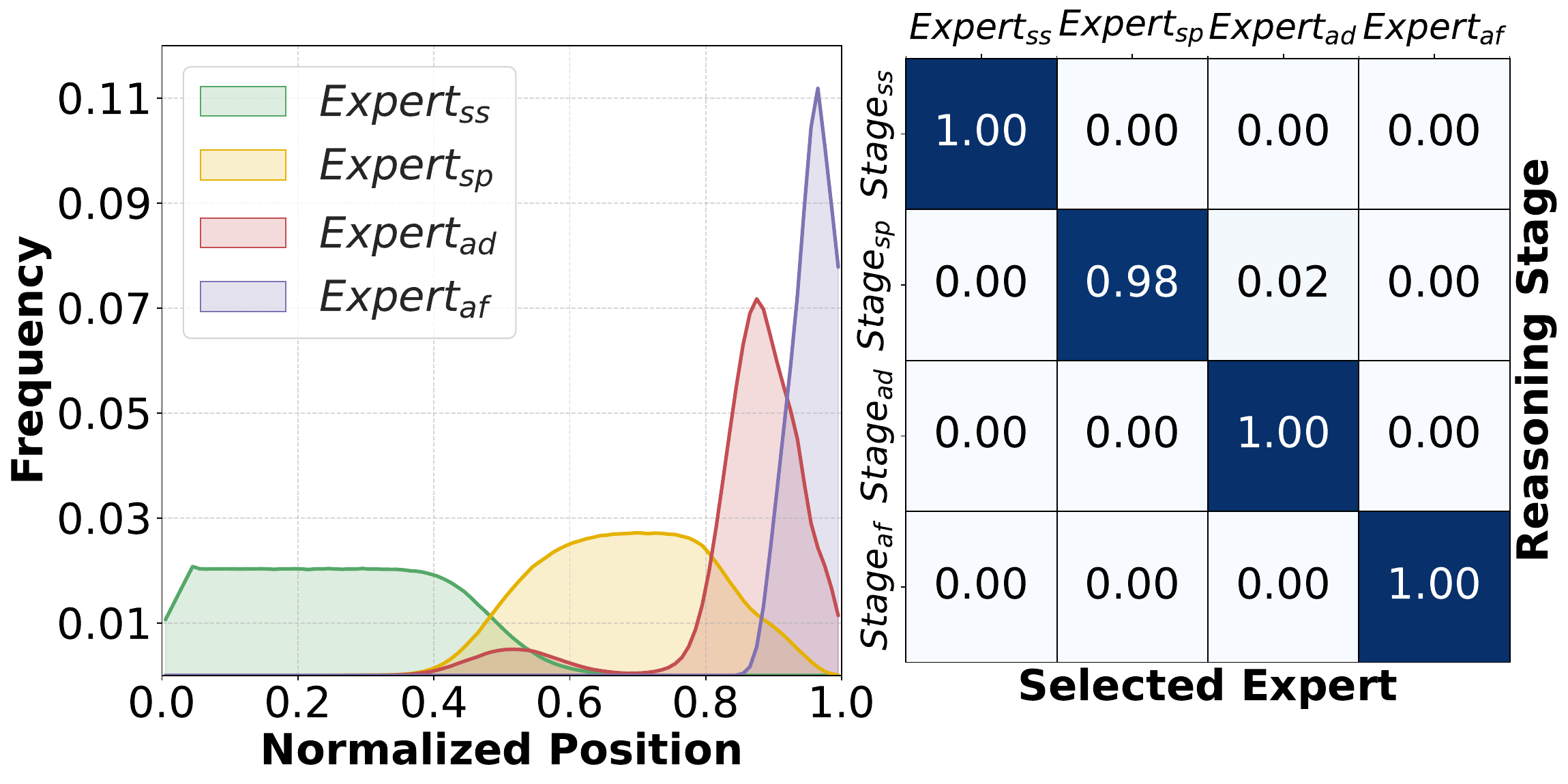}
  \end{subfigure}
  \begin{subfigure}[b]{0.49\textwidth}
    \centering
    \includegraphics[width=\linewidth]{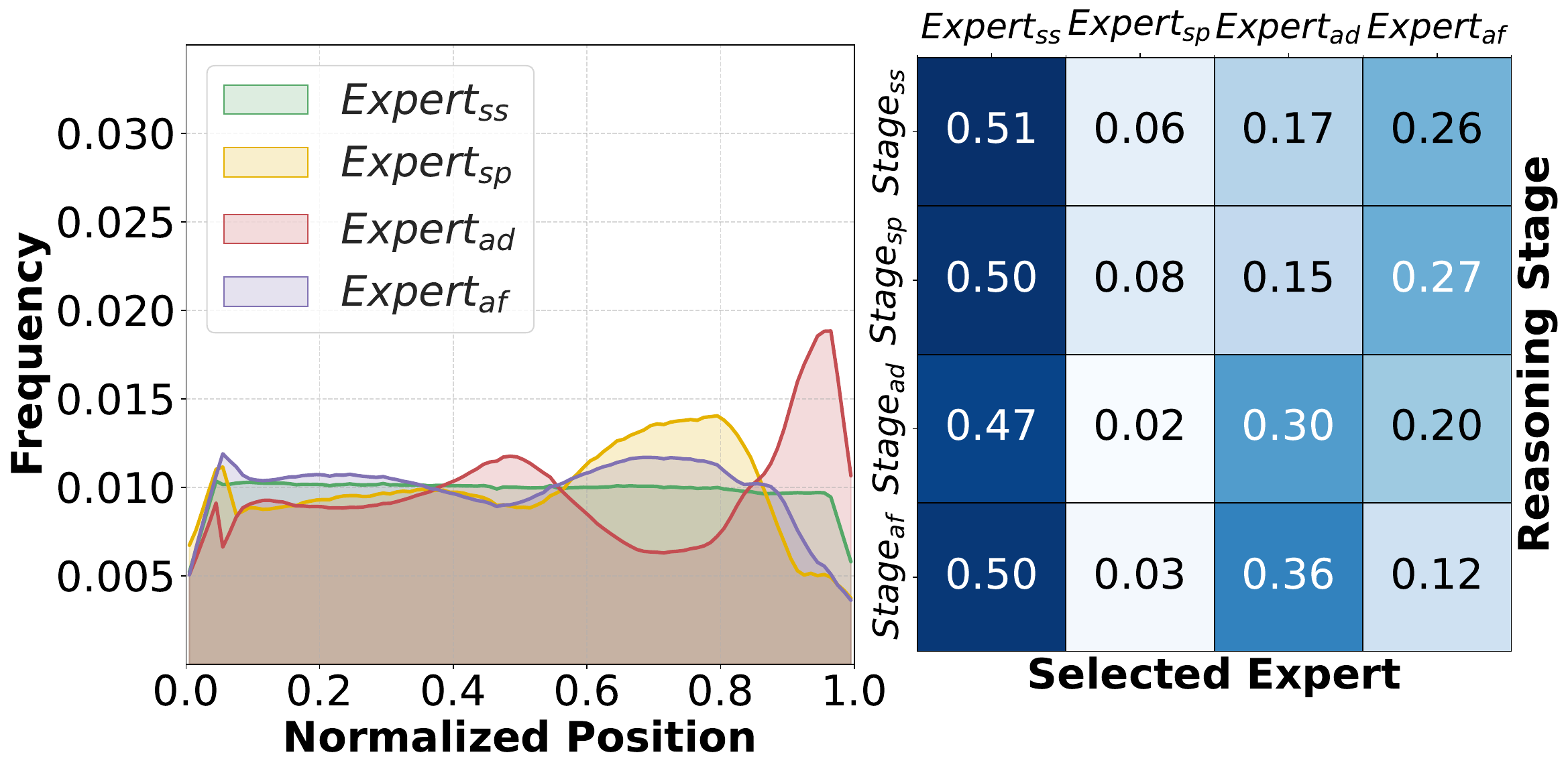}
  \end{subfigure}
  \caption{Expert distribution of \model (left) and MoE (right).}
  \label{fig:expert_distribution}
\vspace{-10pt}
\end{figure*}

\subsubsection{Analysis of \model architecture.}

\textbf{\textit{Q1: How \model performs across different capabilities?}}
\model integrates four specialized experts: screen summary ($\mathcal{E}\text{ss}$), subtask plan ($\mathcal{E}\text{sp}$), action decision ($\mathcal{E}\text{ad}$), and action function ($\mathcal{E}\text{af}$). 
We compare \model with Dense and MoE model, and use specific-capability expert as reference baseline in Figure~\ref{fig:expert_performance}.
Dense and MoE method struggle in later reasoning stages, because the first two stages (screen summary and subtask plan) account for 80\% of the tokens, which dominate CoT training and leave later stages inadequately trained.
Conversely, by activating the expert aligned with reasoning stage, \textbf{\model enables CoT-FT to optimize each expert in its specialized stage, resulting in balanced improvements across all capabilities.}
The superior performance of action-function expert is because of leveraging ground‐truth action decisions, whereas others use their own predicted decisions. By disentangling and fine‐tuning each expert for its specific capability, \model strengthens intermediate reasoning and effectively mitigates error propagation.

\begin{table}[]
\renewcommand\arraystretch{1.0}
\centering
\caption{Performance on AMEX auxiliary GUI tasks.}
\label{tab:amex_aux}
\resizebox{\linewidth}{!}{%
\begin{tabular}{lcccc}
\toprule
\midrule
\textbf{Model} & \textbf{Caption} & \textbf{OCR} & \textbf{Grounding} & \textbf{Action} \\
\midrule
Qwen2VL\ (7B) & \textbf{54.53} & 78.87 & 78.89 & 57.99 \\
Ultars\ (7B)  & 54.08 & 78.87 & 79.11 & 70.33 \\
MoE\ (A3B)   & 52.34 & 78.75 & 77.32 & 64.56 \\
CoME\ (5B)    & 53.89 & \textbf{79.67} & \textbf{79.19} & \textbf{72.61} \\
\midrule
\bottomrule
\end{tabular}
}
\vspace{-15pt}
\end{table}

\textbf{\textit{Q2: Why \model improves on \hcr?}}
We compare \model against MoE by analyzing expert activation distributions and selection accuracy at each stage. As shown in Figure~\ref{fig:expert_distribution}, expert activation in \model exhibits clear stage preference (e.g., $\mathcal{E}_{ss}$ for initial screen summary and $\mathcal{E}_{af}$ for final action function) and achieves 99\% selection accuracy, whereas MoE activation remain uniformly distributed and poorly aligned with reasoning stages. 
The experiment demonstrates that \textbf{\model achieves output-oriented activation in \hcr that successfully activate the expert with corresponding capability required by the reasoning stage.}

\textbf{\textit{Q3: How \model performs on other GUI tasks?}}
\model not only achieves effective \hcr, but can also activate appropriate experts to solve other GUI tasks. We construct three categories of GUI tasks on the AMEX test set—Caption, OCR, and Grounding. Comparisons with the main baselines in Table~\ref{tab:amex_aux} demonstrate that \textbf{\model not only performs best on action prediction through \hcr , but also achieves competitive results on other GUI tasks.}

\textbf{\textit{Q4: Does \model support different architecture?}}
We evaluate a 3-expert variant by removing the screen-summary expert, as well as an 8-expert variant where the number of experts in each stage is scaled up by 2×. Experimental results in Table~\ref{tab:expert-variant} demonstrate that \textbf{\model supports different variant and can be adjusted flexibly}.

\begin{figure*}[t]
\setlength{\abovecaptionskip}{1pt}
\setlength{\belowcaptionskip}{-15pt}
\centering
\includegraphics[width=\linewidth]{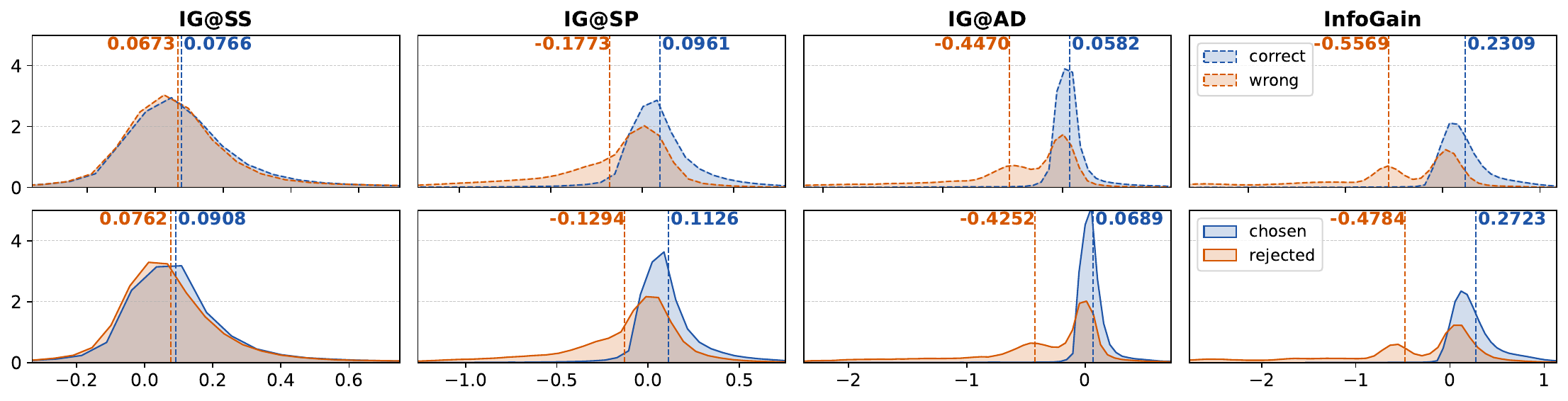}
\caption{Information gain distribution. Above is the sampled data. Below is the DPO data. We illustrate the information gain at screen summary, subtask plan and action decision stage, as well as the total information gain of the entire reasoning trajectory.}
\label{fig:dpo_data}
\end{figure*}

\subsubsection{Analysis of Info-DPO}

\textbf{\textit{Q5: How Info-DPO performs compared with Rule-DPO?}}
We compare Info-DPO with Rule-based reward DPO (Rule-DPO) in Table~\ref{tab:dpo}. 
Experimental results shows that Info-DPO surpasses Rule-DPO on both AITZ (+1.61\%) and AMEX (+2.41\%), and outperforms Rule-DPO across all action categories.
For click actions, Rule-DPO yields marginal overall improvements (+0.51\% and –0.08\%), whereas Info-DPO achieves substantial gains (+3.45\% and +1.70\%). Because click action is highly sensitive to intermediate error: an incorrect element description can lead to large coordinate deviations. Unlike Rule-DPO with no supervision over intermediate reasoning, \textbf{Info-DPO employs information gain to identify and reinforce valid intermediate steps, thereby mitigating error propagation}.

\begin{table}[]
\centering
\caption{Comparison of DPO Strategies.}
\resizebox{\linewidth}{!}{%
\begin{tabular}{lcccccc}
\toprule
\midrule
\textbf{Method} & \textbf{Scroll} & \textbf{Click} & \textbf{Type} & \textbf{Press} & \textbf{Stop} & \textbf{Overall} \\
\midrule
\multicolumn{7}{c}{\cellcolor{bg}\textbf{AITZ}}\\
\midrule
CoT-FT   & 51.74 & 61.77 & 70.00 & 63.44 & 68.84 & 62.25 \\
Rule-DPO & 51.98 & 62.26 & \textbf{86.20} & \textbf{61.80} & 80.36 & 65.37 \\
Info-DPO & \textbf{52.07} & \textbf{65.22} & 83.80 & 59.53 & \textbf{83.33} & \textbf{66.98} \\
\midrule
\multicolumn{7}{c}{\cellcolor{bg}\textbf{AMEX}}\\
\midrule
CoT-FT   & 70.83 & 66.39 & 79.16 & 70.66 & 51.14 & 67.28 \\
Rule-DPO & 82.34 & 66.31 & 78.38 & 48.00 & 58.35 & 70.20 \\
Info-DPO & \textbf{82.43} & \textbf{68.09} & \textbf{82.58} & \textbf{70.67} & \textbf{67.74} & \textbf{72.61} \\
\midrule
\bottomrule
\end{tabular}%
}
\label{tab:dpo}
\vspace{-12pt}
\end{table}

\textbf{\textit{Q6: How InfoGain-reward performs on reasoning trajectory evaluation?}}
Figure~\ref{fig:dpo_data} presents the Information gain distribution for the sampled data and DPO data.
As reasoning progresses, the difference between correct and wrong reasoning increases (0.0766\,/\,0.0673 $\rightarrow$ 0.0582\,/\,-0.4252).
This confirms that the \textbf{InfoGain-reward can evaluate and distinguish the reasoning trajectory, where correct reasoning brings positive InfoGain while wrong reasoning results in negative InfoGain}.
Furthermore, filtering DPO data with InfoGain‐reward raises the information gain of the selected trajectories, demonstrating its effectiveness in removing invalid intermediate steps.

\begin{table}
\centering
\renewcommand\arraystretch{1}
\captionof{table}{Analysis of data pairing strategies.}
\begin{tabular}{lccc}
\toprule
\midrule
\multicolumn{1}{l}{\textbf{Strategy}} &
  \textbf{InfoGain} &
  \textbf{InfoPos} &
  \textbf{Action Acc} \\ \midrule
\multicolumn{1}{l}{cc+cw+lw} &
  -0.50 / 0.26 &
  0.03 / 0.61 &
  77.60 / 65.96 \\
\multicolumn{1}{l}{cc+cw} &
  -0.48 / \textbf{0.27} &
  0.03 / \textbf{0.70} &
  78.60 / \textbf{66.98} \\
\multicolumn{1}{l}{cw+lw} &
  -0.71 / 0.26 &
  0.04 / 0.46 &
  76.88 / 64.07 \\
\multicolumn{1}{l}{cw} &
  -0.71 / 0.27 &
  0.04 / 0.56 &
  77.75 / 65.93 \\
\midrule
\bottomrule
\label{tab:dpo_data_strategy}
\end{tabular}
\vspace{-25pt}
\end{table}

\textbf{\textit{Q7: How to select effective data for Info-DPO?}}
For each action, we sample ten reasoning trajectories and construct DPO pairs using one of three strategies: 
(1) \textbf{cc}: pair two correct trajectories; 
(2) \textbf{cw}: pair a correct trajectory with a wrong one; 
and (3) \textbf{lw}: pair the label with an wrong trajectory. Details in Appendix~\ref{appendix:dpo_pair}. 
Experimental results in Table~\ref{tab:dpo_data_strategy} and Table~\ref{tab:appendix-d} show that, higher InfoGain (informative contribution) and InfoPos 
 (intermediate validity) leads to higher action accuracy.
While, including label reasoning trajectories as chosen outputs (lw) provides no gain because of the distribution gap; introducing suboptimal correct reasoning as rejected outputs (cc) further mitigates invalid intermediate steps. Thus, \textbf{the most effective pairing strategy is to pair trajectories with high information gain as chosen outputs and those exhibiting incorrect outcomes or invalid intermediate steps as rejected outputs}.

\textbf{\textit{Q8: Which reward strategy is better for Info-DPO?}}
We analyse the impact of different reward strategies on Info-DPO, including different model scales and calculation methods.
As shown in Table~\ref{tab:dpo_reward_strategy}, using \textbf{7B reward model can achieve comparable results of 72B reward model and even better}, because better fitness and lower loss of 72B model results in a smaller InfoGain calculated based on the difference, which reduces the distinguishing ability of the reward model. 
Moreover, \textbf{using continuous reward function of 0-1 is better than the discrete reward function of 0/1}, because more fine-grained evaluation is important for click and input action as shown in Table~\ref{tab:appendix-e}.

\begin{table}
\centering
\renewcommand\arraystretch{1}
\setlength{\abovecaptionskip}{5pt}
\captionof{table}{Analysis of reward strategies.}
\begin{tabular}{lccc}
\toprule
\midrule
\multicolumn{1}{l}{\textbf{Method}} & \textbf{InfoGain} & \textbf{InfoPos} & \textbf{Action} \\ \midrule
\multicolumn{1}{l}{Reward-7B}  & 0.9841 & 0.6610 & 78.60 / 66.98 \\
\multicolumn{1}{l}{Reward-72B} & 0.8365 & 0.4533 & 78.52 / 66.85 \\ \midrule
\multicolumn{1}{l}{Continuous} & 0.9841 & 0.6610 & 78.60 / 66.98 \\
\multicolumn{1}{l}{Discrete}   & 0.7541 & 0.5018 & 77.94 / 65.51 \\
\midrule
\bottomrule
\label{tab:dpo_reward_strategy}
\end{tabular}
\vspace{-18pt}
\end{table}

\subsubsection{Analysis of Efficiency}

\begin{table}[]
\centering
\caption{Efficiency analysis of CoME architecture.}
\renewcommand\arraystretch{1}
\begin{tabular}{lcccc}
\toprule
\midrule
\multirow{2}{*}{\textbf{Method}} & \multicolumn{2}{c}{\textbf{GPU Mem (GB)}} & \multicolumn{2}{c}{\textbf{Accuracy (\%)}} \\ \cmidrule(lr){2-3} \cmidrule(lr){4-5}
                                 & \textbf{train}    & \textbf{infer}   & \textbf{type}     & \textbf{match}    \\ \midrule
Qwen2VL\ (7B)                       & 31.62           & 16.52          & 71.59             & 54.46             \\
UITars\ (7B)                       & 32.08           & 16.92          & 78.07             & 65.41             \\
MoE\ (A3B)                      & 30.70           & 22.32          & 72.94             & 60.69             \\
CoME\ (5B)                             & 18.52           & 11.69          & 78.60             & 66.98             \\
\midrule
\bottomrule
\end{tabular}
\label{tab:efficiency}
\vspace{-10pt}
\end{table}

\textbf{\textit{Q9: How efficient is \model?}}
We compare \model with both dense and MoE model in Table~\ref{tab:efficiency}, \textbf{\model achieves better action accuracy while maintaining lower GPU memory usage}. GPU memory usage is dominated by model parameters, while the KV cache during inference contributes only a small fraction. More details in Appendix~\ref{appendix:efficiency}.

\section{Conclusion}

We propose Channel‐of‐Mobile‐Experts (\model), a novel agent architecture implemented with output-oriented activation, to activate the corresponding expert aligned with different stage in \hcr.
We develop a progressive curriculum (Expert-FT, Router-FT, CoT-FT) to achieve decoupled enhancement and balanced integration of different capabilities.
To mitigate error propagation in reasoning, we introduce InfoGain-Driven DPO, which uses information gain to reinforce informative intermediate steps. 
\model performs best on both AITZ and AMEX, and extensive experiments demonstrate the effectiveness of the novel architecture and training strategies.

\newpage

\section{Impact statement}

We introduce hybrid-capabilities reasoning in mobile scenarios, in which a model, given the current state, need to proceed through multiple reasoning stages to arrive at an action decision—each stage potentially requiring a different capability. Such multi-stage reasoning is ubiquitous in agent systems. When an agent decides on an action, it typically engages in environment perception, task planning, and action-function generation, all of which exemplify hybrid-capabilities reasoning. Consequently, the Channel-of-Mobile-Experts architecture proposed in this paper can be applied to a wide range of agent tasks, and our training strategy can be employed to train  diverse agent applications.

\bibliography{come}
\bibliographystyle{icml2026}

\onecolumn
\appendix

\section{Limitation of Mixture-of-Experts on \hcr}\label{appendix:moe}

\begin{figure}[h]
\centering
\includegraphics[width=\linewidth]{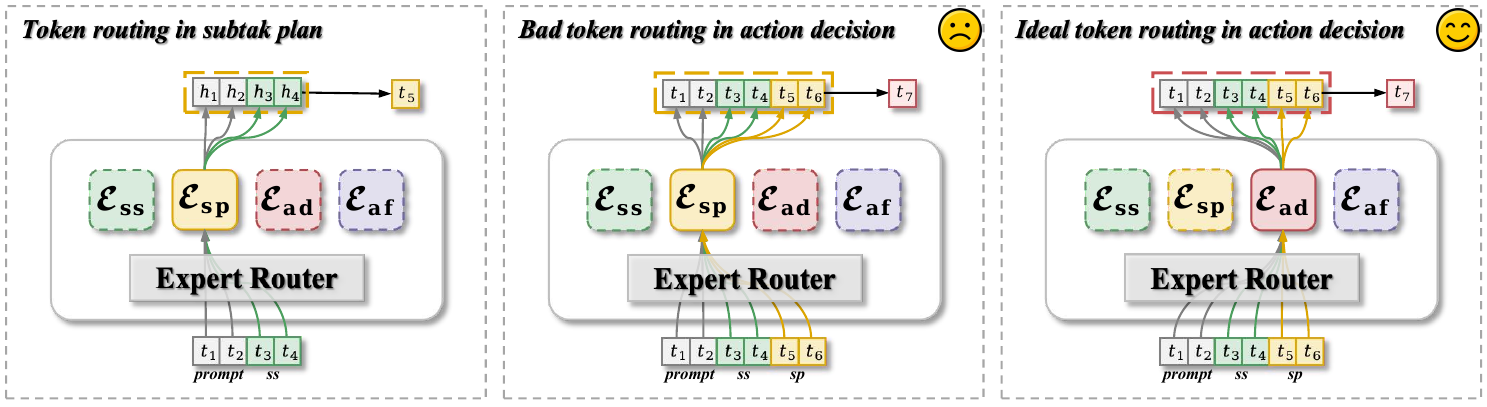}
\caption{A case to show the limitation of MoE on \hcr.}
\label{fig:moe-limitation}
\end{figure}

Mixture-of-Experts (MoE)~\cite{jacobs1991adaptive} architectures have gradually gained traction in LLMs. By introducing multiple FFN modules as experts alongside a routing/gating mechanism, MoE sparsely activates only a subset of the experts for each token in the input sequence, thereby enabling an increase in total parameter while keeping computational costs relatively manageable~\cite{zhou2022mixture,jiang2024mixtral,dai2024deepseekmoe}.
By assigning different subsets in the input to different experts, MoE implicitly links distinct capabilities to some specific experts, thereby achieving capability disentanglement in hybrid‐capability scenarios. However, this represents an input-oriented activation, which is better for handling multi-task or multi-modal inputs. 

Mobile Agents present an even more complex scenario. In order to achieve accurate action prediction, the agent need to perceive the current screen state, plan the next sub-task, then generate the high-level action decision and the low-level action function. This process involves multi-dimensional reasoning capabilities, which is referred to as \hcr. Ideally, we want to activate the experts with the corresponding capability conditioned on the current reasoning stage to generate the output token, hence, this represents an output-oriented activation.

The reason why MoE can not achieve output-oriented activation can be attributed to the auto-regressive mechanism, which forces the same token in the input sequence routed to the same expert when generating tokens in different reasoning stages. As shown in Figure~\ref{fig:moe-limitation}, although we can force the input tokens in prompt and screen summary to be routed to the specialized expert $\mathcal{E}_\text{sp}$ to generate the output token in the subtask plan stage, when generating output tokens in action decision stage, these input tokens will be forwarded to the $\mathcal{E}_\text{sp}$ again instead of the $\mathcal{E}_\text{ad}$. Thus, we can not use $\mathcal{E}_\text{ad}$ to process these preceding tokens, preventing the fully leveraging of capability of $\mathcal{E}_\text{ad}$ in action decision.

\section{DPO data selection}\label{appendix:dpo_pair}

To construct DPO data pairs, we sample $K$ reasoning trajectories $\mathcal{T}$ for each action. These trajectories fall into three categories, and we adopt a tailored pairing strategy for each using the reward $\mathcal{R}_\text{IG}$, $\mathcal{R}_{\text{IG}}^+$, $\mathcal{R}_\text{ACC}$ and $\mathcal{R}_\text{CoT}$:

\textbf{The sampled trajectories are all correct.}
We partition the sampled trajectories by whether $\mathcal{R}_{\text{IG}^+}=1$. 
From the $\mathcal{R}_{\text{IG}^+}=1$ subset, we select the trajectory with the highest $\mathcal{R}_\text{CoT}$ as the chosen output, which represents the most effective reasoning trajectory leading to the correct result. 
Conversely, from the $\mathcal{R}_{\text{IG}^+}=0$ subset, we pick the trajectory with the lowest $\mathcal{R}_\text{IG}$ as the rejected output, to suppress the trajectory that reaches a correct answer while through invalid intermediate reasoning steps.

\begin{equation}
\begin{aligned}
    \text{chosen: }& \arg\max\bigl\{\mathcal{R}_\text{CoT}(T^{(k)})\:|\:T^{(k)}\in\mathcal{T},\mathcal{R}_{\text{IG}}^+(T^{(k)})=1\bigr\}  \\
    \text{rejected: }& \arg\min\bigl\{\mathcal{R}_\text{IG}(T^{(k)})\:|\:T^{(k)}\in\mathcal{T},\mathcal{R}_{\text{IG}}^+(T^{(k)})=0\bigr\}
\end{aligned}
\end{equation}

\textbf{The sampled trajectories are partially correct.}
We partition the sampled trajectories by whether $\mathcal{R}_{\text{ACC}}>0$. 
From the $\mathcal{R}_{\text{ACC}}>0$ subset, we prioritize selecting the trajectory with the highest $\mathcal{R}_\text{CoT}$ and $\mathcal{R}_{\text{IG}^+}=1$ as the chosen output, which represents the most effective reasoning trajectory leading to the correct result. 
From the $\mathcal{R}{\text{ACC}}=0$ subset, we select the trajectory with the lowest $\mathcal{R}{\text{IG}}$ as the rejected output, as it represents the worst reasoning and should be avoided.

\begin{equation}
\begin{aligned}
    \text{chosen: }& 
    \begin{cases}
    \arg\max\bigl\{\mathcal{R}_\text{CoT}(T^{(k)})\:|\:T^{(k)}\in\mathcal{T},\mathcal{R}_{\text{ACC}}^+(T^{(k)})>0,\mathcal{R}_{\text{IG}}^+(T^{(k)})=1\bigr\}\\
    \quad\quad\quad \text{if} \bigl\{T^{(k)}\:|\:\mathcal{R}_{\text{IG}}^+(T^{(k)})=1\big\}\neq\phi \\
    \arg\max\bigl\{\mathcal{R}_\text{CoT}(T^{(k)})\:|\:T^{(k)}\in\mathcal{T},\mathcal{R}_{\text{ACC}}^+(T^{(k)})>0,\mathcal{R}_{\text{IG}}^+(T^{(k)})=0\bigr\}\\
    \quad\quad\quad \text{if} \bigl\{T^{(k)}\:|\:\mathcal{R}_{\text{IG}}^+(T^{(k)})=1\big\}=\phi  \\
    \end{cases} \\
    \text{rejected: }& \arg\min\bigl\{\mathcal{R}_\text{IG}(T^{(k)})\:|\:T^{(k)}\in\mathcal{T},\mathcal{R}_{\text{IG}}^+(T^{(k)})=0\bigr\}
\end{aligned}
\end{equation}

\textbf{The sampled trajectories are all wrong.}
We use the ground truth reasoning trajectory $T^*$ as chosen output and choose the trajectory with the lowest $\mathcal{R}{\text{IG}}$ as the rejected output.

\begin{equation}
\begin{aligned}
    \text{chosen: }& T^*  \\
    \text{rejected: }& \arg\min\bigl\{\mathcal{R}_\text{IG}(T^{(k)})\:|\:T^{(k)}\in\mathcal{T}\bigr\}
\end{aligned}
\end{equation}

\section{Progressive training strategy}

\begin{figure}[h]
\setlength{\belowcaptionskip}{-16pt}
\centering
\includegraphics[width=\linewidth]{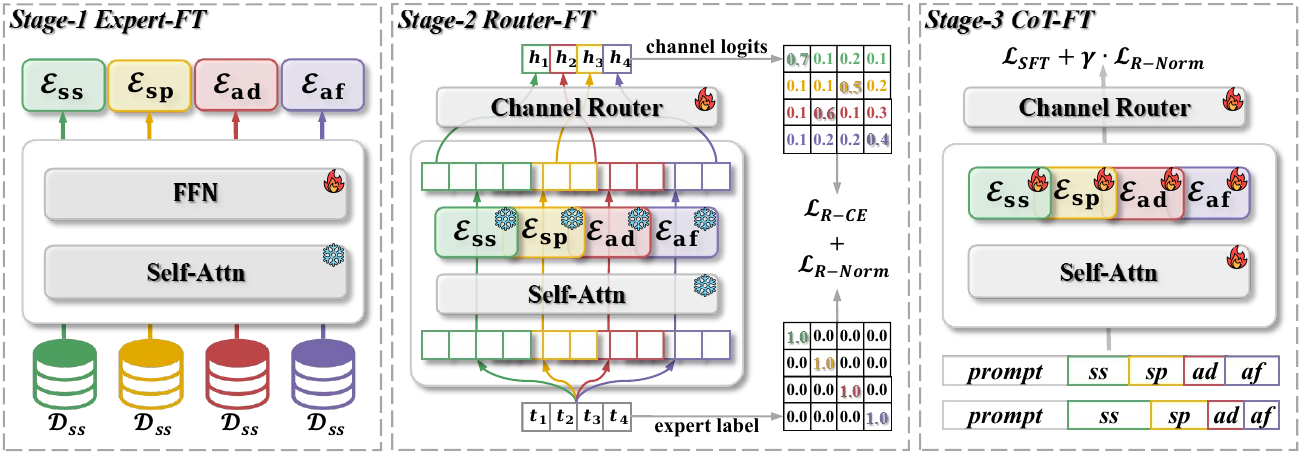}
\caption{Overall training framework of \model.}
\label{fig:come-training}
\end{figure}

In order to empower \model with \hcr, we propose a progressive training strategy consists of three stages: 
(1) \textbf{Expert Finetuning (Expert-FT)}, which explicitly decouples and enhances different capabilities; 
(2) \textbf{Router Finetuning (Router-FT)}, which allows the activation of expert aligned with the current reasoning stage; 
(3) \textbf{Chain-of-Thought Finetuning (CoT-FT)}, which facilitates seamless collaboration and balanced optimization among experts. 
The overall training framework is shown in Figure~\ref{fig:come-training}.

\section{Experiment setup details}\label{appendix:setup}

\subsection{Dataset}

We train and evaluate \model on both AITZ~\cite{zhang2024android} and AMEX~\cite{zhang2024android}.

\textbf{AITZ} is a cleaned subset of the large scale AITW~\cite{rawles2023androidinthewild}, comprising 2.5K unique instructions and 18K steps. These instructions, drawn from over 70 mobile apps, are divided into five categories: general, google apps, install, web shopping, and single. Each step is annotated with a chain-of-action-thought, covering screen summarization, action planning, action description, and action result.

\textbf{AMEX} is a large-scale mobile dataset comprising 104k screenshots and 3k instructions collected from 64 distinct apps. It features approximately 1.6M GUI element grounding annotations and 712k GUI element function descriptions. We further augment AMEX using the data construction methodology established by AITZ, the augmented data will be open-sourced.

\subsection{Baselines}

We compare \model with two types of baselines: (1) Dense mobile agents and (2) Sparse MoE models.

\textbf{Dense mobile agents} are the general mobile agents pre-trained on large mobile datasets with strong capability on mobile environments, including: 

$\bullet$ \textbf{Auto-GUI}~\cite{zhang2024you}
proposes a chain-of-action technique, that incorporates the text described previous action history and future action plan to facilitate action decision on the current screen.

$\bullet$ \textbf{SeeClick}~\cite{cheng2024seeclick}
only relies on screenshots for mobile task automation. SeeClick is pre-trained on large scale of mixed data including widget caption, UI summarization and UI grounding, together with general multi-modal dataset, \eg VQA and visual reasoning.

$\bullet$ \textbf{ShowUI}~\cite{lin2024showui}
uses UI-guide visual token selection to formulate screenshot as an UI connected graph, thus reducing the computational cost. The introduced interleaved vision-language-action streaming can flexibly manage visual-action history in navigation to enhance training efficiency.

$\bullet$ \textbf{UITars}~\cite{qin2025ui}
is a native GUI agent that solely perceives the screenshot as input and perform end-to-end action decision, which incorporates the unified action modeling and system-2 reasoning. UITars is pre-trained on large-scale screenshots with precise caption and grounding annotation. Moreover, it is iterative trained with reflective online traces to continuously learn from its mistakes.

$\bullet$ \textbf{UGround}~\cite{gou2024navigating}
is trained with a large scale of 10M GUI elements and the target bounding box over 1.3M screenshots to empower strong grounding ability on mobile environment.

$\bullet$ \textbf{OS-Atlas}~\cite{wu2024atlas}
is a foundational GUI action model with grounding mode, action mode and agent mode. It can generate an action description as a simple plan and grounding the action on the screenshot, which excels at GUI grounding and OOD agentic tasks.

$\bullet$ \textbf{SphAgent}~\cite{chai2024amex}
is pre-trained on the AMEX dataset, including three-level annotations: interactive element grounding, screen and element functionality descriptions, and instruction with action chains. SphAgent is equipped with strong capabilities on screen understanding and element recognition.

$\bullet$ \textbf{GUI-R1}~\cite{luo2025gui} is a strong reasoning-centric mobile GUI agents. It leverages explicit step-by-step reasoning (R1-style) to plan actions and then executes the predicted operation.

$\bullet$ \textbf{UI-S1}~\cite{lu2025ui} is trained with Semi-online Reinforcement Learning, which simulates online rollouts using offline trajectories to better capture multi-step signals without costly environment interaction.

\textbf{Sparse MoE models} are the MoE base models that are pretrained on the general visual-language tasks and fine-tuned on the mobile environment by us, including:

$\bullet$ \textbf{MolmoE}~\cite{deitke2024molmo}
is a multi-modal MoE MLLM with 1.5B active and 7.2B total parameters, which nearly matches the performance of GPT-4V on both academic benchmarks and human evaluation.

$\bullet$ \textbf{DeepSeekVL2}~\cite{wu2024deepseek}
incorporates a dynamic tiling vision encoding strategy to process high-resolution images and leverages DeepSeekMoE~\cite{dai2024deepseekmoe} with Multi-head Latent Attention and shared experts, which has only 4.5B activated parameters.

$\bullet$ \textbf{AriaUI}~\cite{yang2024aria}
is the first native MoE-based mobile agent that leverage textual or text-image interleaved action from trajectory to enhance the dynamic contexts understanding.

\subsection{Metrics}

Following Auto-GUI~\cite{zhang2024you}, we use \textbf{action type} score to evaluate the accuracy of the predicted action type as well as the \textbf{action match} score to measure the accuracy of the predicted action parameter. For CLICK action, the distance between predicted action and labeled action less than 140 on the normalized 0–1000 coordinate scale is considered correct, or both of the clicked coordinate fall in the same bounding box. For TYPE action, the F1 score between predicted text and labeled text greater than 0.5 is considered correct. For other actions, we use exact match between the predicted action and labeled action.

\subsection{Implementation details}

\model comprises four distinct experts, each initialized from the Feed-Forward Network (FFN) layers of Qwen2VL-2B. During the Expert-FT stage, we exclusively fine-tune the FFN layers while freezing all other parameters. Training is performed on task-specific subsets—screen summary, subtask plan, action decision, and action function—extracted from the original \hcr data of AITZ and AMEX. This stage is conducted using a cosine learning rate scheduler with a peak learning rate of 1e-5. 
In the Router-FT stage, we annotate each output token according to its corresponding reasoning stage and expert. Only the channel router is trained in this stage, using a peak learning rate of 1e-4 with a cosine scheduler.
During the CoT-FT stage, we fine-tune the language model component of \model while freezing the Vision Transformer (ViT). We adopt a LoRA configuration with rank 64 and LoRA alpha 128. This stage uses a peak learning rate of 2e-4 and follows a cosine learning rate schedule.
For Info-DPO, we sample $K=10$ reasoning trajectories for each step. And regarding the thresholds in Eq.~\ref{eq:action_acc}, we set the distance threshold $\delta_d$ to 50 and the function threshold $\delta_f$ to 0.5. During Info-DPO training, we maintain the same configuration as CoT-FT but reduce the peak learning rate to 1e-4.
For the weights of auxiliary SFT loss and Router Norm loss in Eq.~\ref{eq:sft_loss} and Eq.~\ref{eq:dpo_loss}, we set $\alpha$ to 1 and $\beta$ to 0.1.

\section{Analysis of relative improvement results on AITZ}\label{appendix:exp_aitz}

We conducted a more detailed analysis of the performance improvements across all methods. Specifically, we computed the average performance of all methods and treated it as a new baseline (Average). We then measured each method’s improvement relative to this baseline. Additionally, we calculated the variance of accuracy improvements across different action categories. The results are shown in the Table~\ref{tab:aitz-improve}.
Only four methods achieved consistent performance improvements across all action types. Among them, \model demonstrated the highest relative improvement (+11.56\%) while maintaining a lower improvement bias (4.41). Furthermore, we observed that methods incorporating multiple experts, such as MoE and \model, exhibit lower improvement bias across different action categories compared to Dense models. This indicates that leveraging specialized experts not only enhances overall performance but also contributes to greater robustness of the method.

\begin{table*}[h]
\centering
\caption{Detailed results on AITZ with relative improvement and improvement variance.}
\renewcommand\arraystretch{1.2}
\resizebox{0.9\textwidth}{!}{
\begin{tabular}{lccccccc}
\toprule
\midrule
\multicolumn{1}{l}{\textbf{Method}} & \textbf{SCROLL} & \textbf{CLICK} & \textbf{TYPE} & \textbf{PRESS} & \textbf{STOP} & \textbf{OVERALL} & \textbf{IMPROVE} \\ \midrule
Average                              & 43.57           & 51.92          & 74.15         & 57.74          & 69.07         & 55.41            & -            \\ \midrule
AutoUI       & 61.40$_{\textcolor{red}{+17.83}}$ & 32.20$_{\textcolor{softgreen}{-19.72}}$ & 81.40$_{\textcolor{red}{+7.25}}$ & 57.70$_{\textcolor{softgreen}{-0.04}}$ & 74.40$_{\textcolor{red}{+5.33}}$ & 47.69 & -7.73$_{12.37}$ \\
Qwen2VL(2B)      & 23.05$_{\textcolor{softgreen}{-20.52}}$ & 43.74$_{\textcolor{softgreen}{-8.18}}$ & 59.40$_{\textcolor{softgreen}{-14.75}}$ & 52.22$_{\textcolor{softgreen}{-5.52}}$ & 47.02$_{\textcolor{softgreen}{-22.05}}$ & 43.80 & -11.62$_{6.53}$ \\
ShowUI       & 23.79$_{\textcolor{softgreen}{-19.78}}$ & 40.34$_{\textcolor{softgreen}{-11.58}}$ & 85.20$_{\textcolor{red}{+11.05}}$ & 59.00$_{\textcolor{red}{+1.26}}$ & 87.55$_{\textcolor{red}{+18.48}}$ & 49.55 & -5.87$_{14.07}$ \\
UITars(2B)       & 30.17$_{\textcolor{softgreen}{-13.40}}$ & 55.25$_{\textcolor{red}{+3.33}}$ & 84.77$_{\textcolor{red}{+10.62}}$ & 53.78$_{\textcolor{softgreen}{-3.96}}$ & 61.39$_{\textcolor{softgreen}{-7.68}}$ & 55.63 & +0.21$_{8.40}$ \\
SeeClick     & 11.14$_{\textcolor{softgreen}{-32.43}}$ & 52.96$_{\textcolor{red}{+1.04}}$ & 53.00$_{\textcolor{softgreen}{-21.15}}$ & 67.88$_{\textcolor{red}{+10.14}}$ & 55.36$_{\textcolor{softgreen}{-13.71}}$ & 49.11 & -6.31$_{15.23}$ \\
Qwen2VL(7B)      & 43.28$_{\textcolor{softgreen}{-0.29}}$ & 55.30$_{\textcolor{red}{+3.38}}$ & 51.40$_{\textcolor{softgreen}{-22.75}}$ & 56.92$_{\textcolor{softgreen}{-0.82}}$ & 64.87$_{\textcolor{softgreen}{-4.20}}$ & 54.46 & -0.96$_{9.22}$ \\
UIGround     & 58.22$_{\textcolor{red}{+14.65}}$ & 58.48$_{\textcolor{red}{+6.56}}$ & 73.85$_{\textcolor{softgreen}{-0.30}}$ & 58.22$_{\textcolor{red}{+0.48}}$ & 68.78$_{\textcolor{softgreen}{-0.29}}$ & 60.19 & +4.77$_{5.81}$ \\
OS-Atlas     & 76.12$_{\textcolor{red}{+32.55}}$ & 54.83$_{\textcolor{red}{+2.91}}$ & 88.60$_{\textcolor{red}{+14.45}}$ & 68.67$_{\textcolor{red}{+10.93}}$ & 81.75$_{\textcolor{red}{+12.68}}$ & 65.11 & +9.69$_{9.75}$ \\
UITars(7B)       & 56.50$_{\textcolor{red}{+12.93}}$ & 63.87$_{\textcolor{red}{+11.95}}$ & 85.77$_{\textcolor{red}{+11.62}}$ & 58.22$_{\textcolor{red}{+0.48}}$ & 71.29$_{\textcolor{red}{+2.22}}$ & 65.41 & +9.99$_{5.34}$ \\
MolmoE       & 28.19$_{\textcolor{softgreen}{-15.38}}$ & 33.17$_{\textcolor{softgreen}{-18.75}}$ & 62.00$_{\textcolor{softgreen}{-12.15}}$ & 37.07$_{\textcolor{softgreen}{-20.67}}$ & 54.96$_{\textcolor{softgreen}{-14.11}}$ & 38.23 & -17.19$_{3.95}$ \\
Qwen2VL-MoE  & 48.75$_{\textcolor{red}{+5.18}}$ & 59.43$_{\textcolor{red}{+7.51}}$ & 74.00$_{\textcolor{softgreen}{-0.15}}$ & 57.70$_{\textcolor{softgreen}{-0.04}}$ & 70.83$_{\textcolor{red}{+1.76}}$ & 60.69 & +5.27$_{3.02}$ \\
AriaUI       & 53.73$_{\textcolor{red}{+10.16}}$ & 60.20$_{\textcolor{red}{+8.28}}$ & 80.80$_{\textcolor{red}{+6.65}}$ & 63.70$_{\textcolor{red}{+5.96}}$ & 76.38$_{\textcolor{red}{+7.31}}$ & 63.56 & +8.14$_{1.46}$ \\
\model  & 52.07$_{\textcolor{red}{+8.50}}$ & 65.22$_{\textcolor{red}{+13.30}}$ & 83.80$_{\textcolor{red}{+9.65}}$ & 59.53$_{\textcolor{red}{+1.79}}$ & 83.33$_{\textcolor{red}{+14.26}}$ & 66.98 & +11.56$_{4.41}$ \\
\midrule
\bottomrule
\end{tabular}
}
\label{tab:aitz-improve}
\end{table*}

\section{Analysis of \model architecture flexibility}

\model decouples diverse capabilities by assigning them to different experts, and integrates these capabilities during reasoning via output-oriented activation. \model can flexibly adjust its architecture according to the capabilities required by a task, and can also scale up the model by increasing the number of experts for each capability.
To validate the flexibility of the \model architecture, we remove the screen-summary expert to create a 3-expert variant and evaluate performance under fewer reasoning stages. To further study scalability, we scale the number of experts to 2x at each stage, resulting in an 8-expert variant.
We compare variants with different numbers of experts during the CoT-SFT stage on AITZ dataset, and report the results in the Table~\ref{tab:expert-variant}.
The results demonstrate that \model supports architectures with different numbers of experts, and that action accuracy improves as the number of experts increases.

\begin{table*}[h]
\renewcommand\arraystretch{1}
\centering
\caption{Comparison of \model variants with different numbers of experts. }
\resizebox{0.85\textwidth}{!}{
\begin{tabular}{lcccccccccc}
\toprule
\midrule
\multirow{2}{*}{\textbf{Method}} & \multirow{2}{*}{\textbf{\#Expert}} & \multirow{2}{*}{\textbf{SCROLL}} & \multicolumn{2}{c}{\textbf{CLICK}} & \multicolumn{2}{c}{\textbf{TYPE}} & \multirow{2}{*}{\textbf{PRESS}} & \multirow{2}{*}{\textbf{STOP}} & \multicolumn{2}{c}{\textbf{Overall}} \\ \cmidrule(lr){4-5} \cmidrule(lr){6-7} \cmidrule(lr){10-11}
    &  &   & \textbf{type}   & \textbf{match}   & \textbf{type}   & \textbf{match}  &  & & \textbf{type}   & \textbf{match}  \\ \midrule
\multirow{3}{*}{\textbf{\model}} & 3-expert   & 41.22 & 81.44 & 61.52 & 83.40 & 79.80 & 66.57 & 71.31 & 74.42 & 62.28 \\
                & 4-expert   & 52.31 & 81.90 & 62.42 & 75.25 & 71.80 & 63.37 & 69.24 & 75.10 & 62.93 \\
                & 8-expert   & 54.39 & 81.12 & 61.96 & 83.60 & 79.20 & 56.92 & 73.02 & 75.66 & 63.59 \\
\midrule
\bottomrule
\end{tabular}
}
\label{tab:expert-variant}
\end{table*}

\section{Analysis of DPO data selection strategies}\label{appendix:exp_dpo_data}

For each action, we sample ten reasoning trajectories and construct DPO pairs using one of three strategies as described in Appendix~\ref{appendix:dpo_pair}: 
(1) \textbf{cc}: pair two correct trajectories, if the sampled trajectories are all correct; 
(2) \textbf{cw}: pair a correct trajectory with a wrong one, if the sampled trajectories are partially correct; 
and (3) \textbf{lw}: pair the label with an wrong trajectory, if the sampled trajectories are all wrong.
We provide more detailed analysis of information gain comparison among different combination of the pairing strategy as shown in Table~\ref{tab:appendix-d}. Higher InfoGain and InfoPositive will lead to higher action accuracy, because the higher InfoGain indicates that the reasoning trajectory provides more useful information, and higher InfoPositive shows that the intermediate reasoning step is much more valid. 
Introducing pair data from lw will decrease the InfoGain in reject output, because lw shows that the task is difficult for \model to finish, the trajectory with the lowest $\mathcal{R}_\text{IG}$ contains less useful information. Introducing pair data from cc increase the InfoGain in reject output, because some suboptimal correct trajectories are added to the rejected output.

\vspace{-5pt}
\begin{table*}[h]
\centering
\caption{Analysis of data selection strategy}
\renewcommand\arraystretch{1.2}
\resizebox{\textwidth}{!}{%
\begin{tabular}{lcccccccccccc}
\toprule
\midrule
\multicolumn{1}{l}{\multirow{2}{*}{\textbf{Strategy}}} &
  \multicolumn{2}{c}{\textbf{IG@SS}} &
  \multicolumn{2}{c}{\textbf{IG@AP}} &
  \multicolumn{2}{c}{\textbf{IG@AD}} &
  \multicolumn{2}{c}{\textbf{InfoGain}} &
  \multicolumn{2}{c}{\textbf{InfoPositive  }} &
  \multicolumn{2}{c}{\textbf{Action Accuracy}} \\ \cmidrule(lr){2-3} \cmidrule(lr){4-5} \cmidrule(lr){6-7} \cmidrule(lr){8-9} \cmidrule(lr){10-11} \cmidrule(lr){12-13}
\multicolumn{1}{l}{} &
  \textbf{reject} &
  \multicolumn{1}{c}{\textbf{choose}} &
  \textbf{reject} &
  \multicolumn{1}{c}{\textbf{choose}} &
  \textbf{reject} &
  \multicolumn{1}{c}{\textbf{choose}} &
  \textbf{reject} &
  \multicolumn{1}{c}{\textbf{choose}} &
  \textbf{reject} &
  \multicolumn{1}{c}{\textbf{choose}} &
  \textbf{type} &
  \textbf{match} \\ \midrule
\multicolumn{13}{c}{\cellcolor{bg}\textbf{AITZ}} \\ \midrule
\multicolumn{1}{l}{cc+cw+lw} &
  0.0748 &
  \multicolumn{1}{c}{0.0234} &
  -0.1430 &
  \multicolumn{1}{c}{0.1595} &
  -0.4354 &
  \multicolumn{1}{c}{0.0814} &
  -0.5037 &
  \multicolumn{1}{c}{0.2645} &
  \underline{0.0291} &
  \multicolumn{1}{c}{0.6115} &
  77.60 &
  65.96 \\
\multicolumn{1}{l}{cc+cw} &
  0.0762 &
  \multicolumn{1}{c}{\textbf{0.0908}} &
  -0.1294 &
  \multicolumn{1}{c}{0.1125} &
  -0.4252 &
  \multicolumn{1}{c}{0.0689} &
  -0.4784 &
  \multicolumn{1}{c}{\textbf{0.2723}} &
  0.0300 &
  \multicolumn{1}{c}{\textbf{0.7045}} &
  78.60 &
  66.98 \\
\multicolumn{1}{l}{cw+lw} &
  \underline{0.0686} &
  \multicolumn{1}{c}{-0.0103} &
  \underline{-0.2049} &
  \multicolumn{1}{c}{\textbf{0.1790}} &
  -0.5758 &
  \multicolumn{1}{c}{\textbf{0.0905}} &
  -0.7121 &
  \multicolumn{1}{c}{0.2592} &
  0.0380 &
  \multicolumn{1}{c}{0.4631} &
  76.88 &
  64.07 \\
\multicolumn{1}{l}{cw} &
  0.0695 &
  \multicolumn{1}{c}{0.0808} &
  -0.1973 &
  \multicolumn{1}{c}{0.1143} &
  \underline{-0.5869} &
  \multicolumn{1}{c}{0.0741} &
  \underline{-0.7147} &
  \multicolumn{1}{c}{0.2693} &
  0.0409 &
  \multicolumn{1}{c}{0.5654} &
  77.75 &
  65.93 \\ \midrule
\multicolumn{13}{c}{\cellcolor{bg}\textbf{AMEX}} \\ \midrule
\multicolumn{1}{l}{cc+cw+lw} &
  0.0178 &
  \multicolumn{1}{c}{0.0368} &
  -0.1735 &
  \multicolumn{1}{c}{\textbf{0.1373}} &
  -0.3803 &
  \multicolumn{1}{c}{0.0327} &
  -0.5360 &
  \multicolumn{1}{c}{0.2068} &
  \underline{0.0159} &
  \multicolumn{1}{c}{0.4824} &
  83.79 &
  70.04 \\
\multicolumn{1}{l}{cc+cw} &
  0.0175 &
  \multicolumn{1}{c}{\textbf{0.0385}} &
  -0.1629 &
  \multicolumn{1}{c}{0.1359} &
  -0.3630 &
  \multicolumn{1}{c}{0.0337} &
  -0.5085 &
  \multicolumn{1}{c}{\textbf{0.2083}} &
  0.0172 &
  \multicolumn{1}{c}{\textbf{0.5111}} &
  84.46 &
  72.61 \\
\multicolumn{1}{l}{cw+lw} &
  0.0054 &
  \multicolumn{1}{c}{0.0172} &
  \underline{-0.2268} &
  \multicolumn{1}{c}{0.1293} &
  \underline{-0.4257} &
  \multicolumn{1}{c}{0.0342} &
  \underline{-0.6470} &
  \multicolumn{1}{c}{0.1808} &
  0.0183 &
  \multicolumn{1}{c}{0.3797} &
  84.44 &
  71.14 \\
\multicolumn{1}{l}{cw} &
  \underline{0.0038} &
  \multicolumn{1}{c}{0.0166} &
  -0.2199 &
  \multicolumn{1}{c}{0.1266} &
  -0.4103 &
  \multicolumn{1}{c}{\textbf{0.0357}} &
  -0.6265 &
  \multicolumn{1}{c}{0.1790} &
  0.0199 &
  \multicolumn{1}{c}{0.4004} &
  84.92 &
  72.15 \\
\midrule
\bottomrule
\end{tabular}%
}
\label{tab:appendix-d}
\vspace{-10pt}
\end{table*}

\section{Analysis of reward strategies}\label{appendix:exp_dpo_strategy}

We provide detailed analysis of different strategies in Info-DPO, including different reward model scales and reward function design in Table~\ref{tab:appendix-e}. As for the reward model scale, using 7B reward model can achieve comparable and even better overall performance compared with 72B reward model. While 72B reward model performs much better on CLICK action, because click action is much more challenging that requires precise description of the element to click and coordinates of the element grounding. However, 72B reward results in lower InfoGain which is highly relevant to the action accuracy, because 72B reward model have a better fitness and lower cross-entropy loss, thus the difference of information entropy between different stage is much more smaller.  
As for different reward function designs, the continuous reward function proves to be more effective, particularly on the CLICK action. This is because a smaller distance between the predicted and ground-truth coordinates indicates higher prediction accuracy, and such outputs should be more likely selected as the chosen output for model optimization. However, this level of granularity cannot be captured by discrete reward functions.

\begin{table*}[h]
\centering
\caption{Analysis of reward strategy}
\renewcommand\arraystretch{1.1}
\resizebox{0.9\textwidth}{!}{%
\begin{tabular}{lcccccccccc}
\toprule
\midrule
\multicolumn{1}{l}{\multirow{2}{*}{\textbf{Method}}}           & \multicolumn{6}{c}{\textbf{ACTION ACCURACY}}                                  & \multicolumn{4}{c}{\textbf{REWARD MARGIN}}    \\ \cmidrule(lr){2-7} \cmidrule(lr){8-11}
\multicolumn{1}{l}{}           & \textbf{SCROLL} & \textbf{CLICK} & \textbf{TYPE}  & \textbf{PRESS} & \textbf{STOP}  & \multicolumn{1}{c}{\textbf{OVERALL}} & \textbf{IG@SS} & \textbf{IG@AP}   & \textbf{IG@AD} & \textbf{InfoGain} \\ \midrule
\multicolumn{11}{c}{\cellcolor{bg}\textbf{Reward Model Scale}}                                                                                                        \\ \midrule
\multicolumn{1}{l}{Reward-7B}         & 52.07  & 65.22 & 83.80 & 59.53 & 83.33 & \multicolumn{1}{c}{66.98}   & 0.0113 & 0.3117 & 0.6610 & 0.9841  \\
\multicolumn{1}{l}{Reward-72B}        & 53.06  & 67.37 & 82.20 & 54.57 & 74.60 & \multicolumn{1}{c}{66.85}   & 0.0087 & 0.1681 & 0.6597 & 0.8365  \\ \midrule
\multicolumn{11}{c}{\cellcolor{bg}\textbf{Reward Function Design}}                                                                                                  \\ \midrule
\multicolumn{1}{l}{Continuous} & 52.07  & 65.22 & 83.80 & 59.53 & 83.33 & \multicolumn{1}{c}{66.98}   & 0.0113 & 0.3117 & 0.6610 & 0.9841  \\
\multicolumn{1}{l}{Discrete}   & 54.71  & 62.73 & 83.40 & 61.36 & 78.05 & \multicolumn{1}{c}{65.51}   & 0.0143 & 0.2379 & 0.5018 & 0.7541  \\
\midrule
\bottomrule
\end{tabular}%
}
\label{tab:appendix-e}
\end{table*}
\vspace{-12pt}

\section{Analysis of training reward model}

InfoGain reward is estimated based on the difference in information entropy before and after introducing a specific reasoning stage. This estimation leverages a capability that language models naturally possess. Ideally, any logically consistent reasoning should make it easier for a model to predict the correct action. Therefore, we compare the effectiveness of training-based and training-free reward models, as shown in the Table~\ref{tab:appendix-h}. Training-free reward model could achieve comparable performance with training-based one, indicating that the estimation of InfoGain reward is a inner capability of the LLM with the strong general language modeling ability. Experimental results also demonstrate that InfoGain reward is a robust method to evaluate the CoT reasoning process. 

\begin{table}[h]
\centering
\caption{Comparison of training reward model}
\renewcommand\arraystretch{1.1}
\resizebox{0.4\textwidth}{!}{%
\begin{tabular}{lcc}
\toprule
\midrule
\textbf{Reward Model} & \textbf{Reward Acc} & \textbf{Action Acc} \\ \midrule
training-free         & 82.24               & 66.64               \\
training-based        & 84.68               & 66.98               \\
\midrule
\bottomrule
\end{tabular}%
}
\label{tab:appendix-h}
\end{table}

\section{Analysis of efficiency}\label{appendix:efficiency}

\subsection{Analysis of GPU memory consumption}

In this section, we present a more comprehensive analysis of GPU memory consumption during inference. Specifically, we compare the inference computational cost and action accuracy across three representative architectures: a 7B dense model, a 2B×8 MoE model, and our proposed \model. The results, summarized in Table~\ref{tab:appendix-i}, demonstrate that \model achieves the best overall trade-off, delivering the highest action accuracy while maintaining the lowest GPU memory footprint.
The key memory advantage of \model lies in its architecture design: the parameter set dominates the overall memory usage, whereas the additional KV cache overhead incurred during inference remains relatively minor. Taken together, these results underscore that \model not only achieves best accuracy but also remains resource-efficient.

\begin{table}[h]
\centering
\caption{Detailed analysis of GPU memory usage}
\renewcommand\arraystretch{1.1}
\resizebox{0.5\textwidth}{!}{%
\begin{tabular}{lcccc}
\toprule
\midrule
            & \textbf{Model} & \textbf{Extra} & \textbf{Total} & \textbf{Acc} \\ \midrule
Qwen2VL\ (7B)  & 15.48~GB          & 1.04~GB           & 16.52~GB          & 54.46        \\
{UITars\ (7B)}      & {15.49~GB}          & {1.43~GB}           & {16.92~GB}          & {65.41}        \\
MoE\ (A3B) & 21.34~GB          & 0.98~GB           & 22.32~GB          & 60.69        \\
CoME\ (5B)        & 10.55~GB          & 1.14~GB           & 11.69~GB          & 66.98        \\
\midrule
\bottomrule
\end{tabular}%
}
\label{tab:appendix-i}
\end{table}

\subsection{Analysis of computational cost}

In this section, we provide a theoretical analysis of the computational cost. We compare the FLOPs of \model with strong baseline UITars at each layer. We denote $H$ as the hidden size, $L$ as the sequence length, and $I$ as the FFN's intermediate size of \model. From the model's configuration files, we can find that:

\begin{equation}
    H_{\text{UITars}} \approx  2.33 H, \quad I_{\text{UITars}} \approx 2.11 I
\notag
\end{equation}

In the standard transformer layer, the computation cost of attention layer and FFN layer can be estimated by:

\begin{equation}
    \begin{aligned}
        \text{Cost}_{\text{Attn}} & = 4LH^2+2L^2H \\
        \text{Cost}_{\text{FFN}} &= 2LHI
    \end{aligned}
\notag
\end{equation}

Thus, for \model, we forward the hidden states from different channels in parallel to the attention layer and the expert layer:

\begin{equation}
\begin{aligned}
\text{Cost}_{\text{CoME}} &= 4 \times (4LH^2+2L^2H) + 4 \times 2LHI \\
                          &= 16 LH^2 + 8 L^2H + 8 LHI
\end{aligned}
\notag
\end{equation}

For UITars:

\begin{equation}
\begin{aligned}
\text{Cost}_{\text{UITars}} &= 4 \times (4L(2.33H)^2+2L^2(2.33H)) + 4 \times 2L(2.33H)(2.11I) \\
                          &= 21.71 LH^2 + 4.66 L^2H + 9.83 LHI
\end{aligned}
\notag
\end{equation}

This result shows that \model has lower computational cost than UITars at the FFN layer. While at the attention layer, as the sequence length in our settings is smaller than the hidden size, the first item is dominant and \model also has lower computational cost at attention layer.

\newpage
\section{Case Studies}

\subsection{Case study of DPO data pair}

\begin{figure}[h]
\centering
\includegraphics[width=\linewidth]{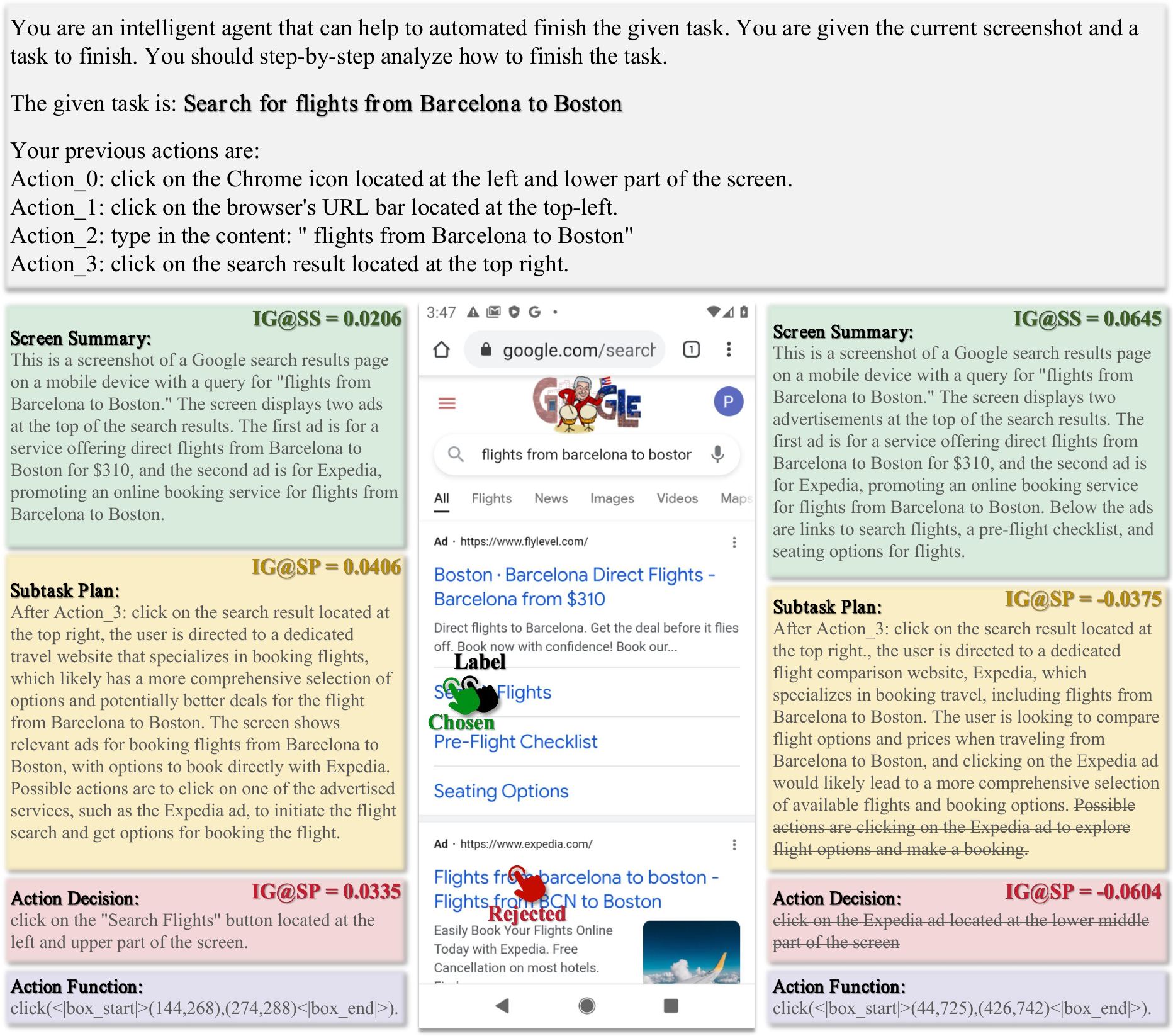}
\caption{InfoGain of DPO data pair where the correct one is chosen and the wrong one is rejected.}
\label{fig:case_infodpo_cw}
\end{figure}

In Figure~\ref{fig:case_infodpo_cw}, we present the InfoGain values at each stage where the correct output is chosen and the wrong output is rejected. It can be observed that the chosen output exhibit InfoGain greater than zero at each stage, indicating that the reasoning in each stage contributes positively to predicting the correct action. For the rejected outputs, however, during the subtask planning stage the model produced an incorrect next action plan, resulting in negative InfoGain. This shows that the reasoning at this stage has a detrimental impact on predicting the correct action. This case illustrates that InfoGain can be used to evaluate the contribution of intermediate reasoning steps to the prediction of the correct action.

\begin{figure}[h]
\centering
\includegraphics[width=\linewidth]{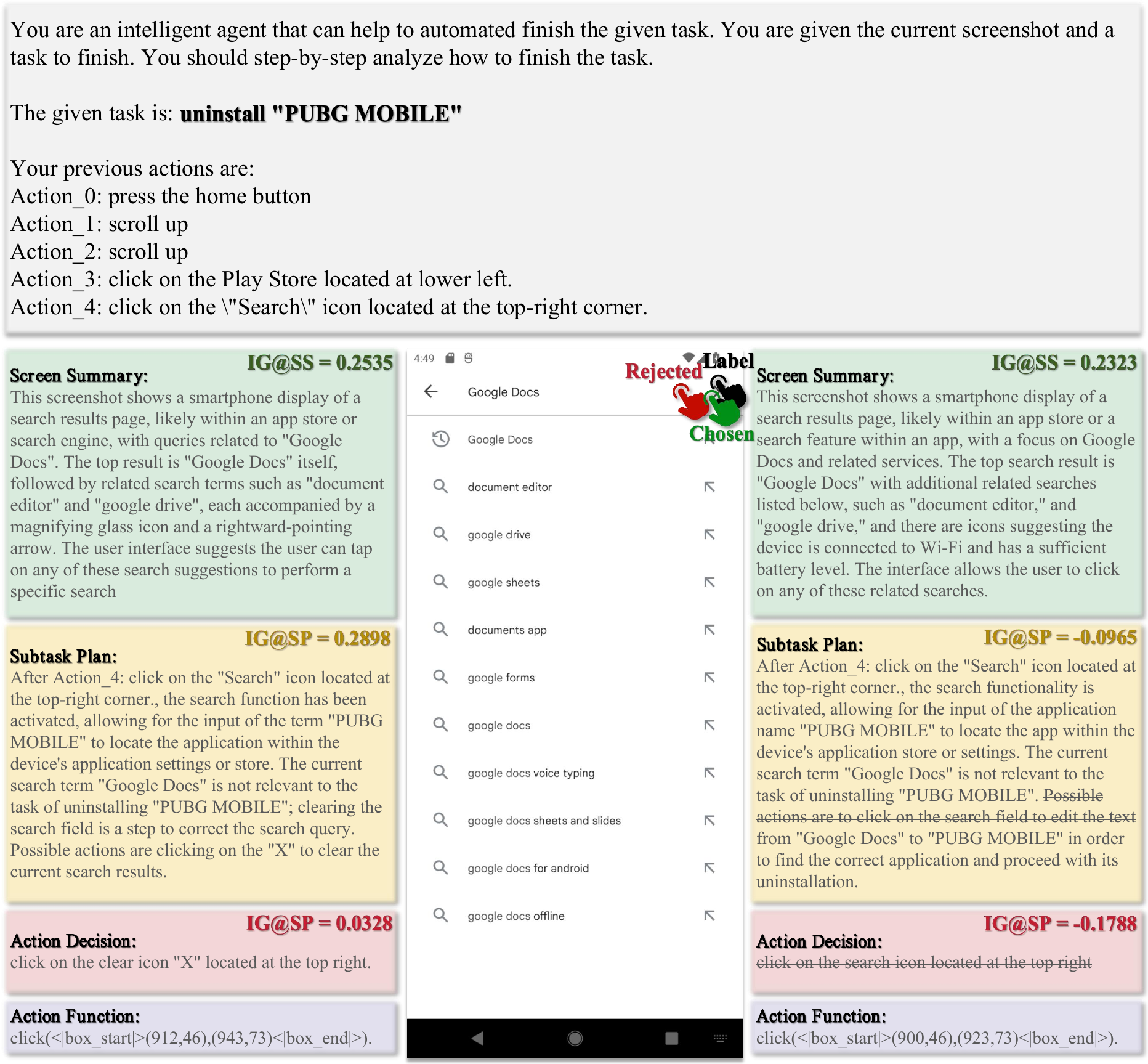}
\caption{InfoGain of DPO data pair where both the chosen and rejected outputs are correct.}
\label{fig:case_infodpo_cc}
\end{figure}

\newpage
In Figure~\ref{fig:case_infodpo_cc}, we present the InfoGain values at each stage where both the chosen and rejected outputs are correct. 
Although the rejected output also result in the correct action, it encounters some reasoning mistakes in subtask plan and action decision stage. Thus the InfoGain at these two stages are negative, indicating that the reasoning process in that stage may be lead to the wrong action. This case illustrates that InfoGain can effectively identify the incorrect intermediate reasoning steps thus suppressing these ineffective reasoning pattern. While this is impossible for the outcome-level reward used by naive DPO.

\end{document}